\documentclass[preprints,article,accept,moreauthors,pdftex]{Definitions/mdpi} 

\firstpage{1} 
\pubvolume{1}
\issuenum{1}
\articlenumber{0}
\pubyear{2021}
\copyrightyear{2021}
\externaleditor{Academic Editor: {Petros Daras}} % For journal Automation, please change Academic Editor to "Communicated by"%please add
\datereceived{28 April 2021 } 
\dateaccepted{17 May 2021} 
\datepublished{} 
\hreflink{https://doi.org/10.3390/s21103569} % If needed use \linebreak

\pdfoutput=1

\usepackage{easyReview}
\setreviewsoff
\usepackage{caption}
\usepackage[labelformat=simple]{subcaption}						
								
\DeclareCaptionLabelFormat{subcaptionlabel}{\normalfont(\textbf{#2}\normalfont)}								
\usepackage{pgf}
\usepackage{ifdraft}
\usepackage{siunitx}
\DeclareSIUnit \year{yr}
\Title{Evaluating the Single-Shot MultiBox Detector %please note the correction to the title
and YOLO Deep Learning Models for the Detection of Tomatoes in a~Greenhouse}
\TitleCitation{Evaluating%please note the correction to the title
 the%please note the correction to the title
 Single-%please note the correction to the title
Shot MultiBox Detector %please note the correction to the title
and YOLO Deep Learning Models for the%please note the correction to the title
 Detection of Tomatoes in a~Greenhouse}
\Author{{Sandro Augusto Magalhães} $^{1,2,}$*\orcidA{}, {Luís Castro} $^{1,2}$\orcidA{}, {Germano Moreira} $^{3}$\orcidC{},{ Filipe Neves dos Santos} $^{1,}$*\orcidD{}, Mário~Cunha $^{1,3}$\orcidE{}, Jorge Dias $^{4}$\orcidF{} and António Paulo Moreira $^{1,2}$\orcidG{}}
\AuthorNames{Sandro Magalhães, Luís Castro, Germano Moreira, Filipe Neves dos Santos, Mário Cunha, Jorge Dias, António Paulo Moreira}
\AuthorCitation{Magalhães, S.; Castro, L.; Moreira, G.; dos Santos, F.N.; Cunha, M.; Dias, J.; Moreira, A.P.}
\address{%
$^{1}$ \quad INESC TEC-Instituto de Engenharia de Sistemas e Computadores, Tecnologia e Ciência, Campus da FEUP, Rua Dr. Roberto Frias, s/n, 4200-465 Porto, Portugal; luis.r.castro@inesctec.pt (L.C.); mccunha@fc.up.pt (M.C.); amoreira@fe.up.pt~(A.P.M.) \\
$^{2}$ \quad Faculty of Engineering, University of Porto, Rua Dr. Roberto Frias, s/n, 4200-465, Porto, Portugal\\
$^{3}$ \quad Faculty of Sciences, University of Porto, Rua do Campo Alegre, s/n, 4169-007 Porto, Portugal\\
$^{4}$ \quad Khalifa University Center for Autonomous Robotic Systems (KUCARS), Khalifa University of Science and
Technology (KU), Abu Dhabi 127788, United Arab Emirates; jorge.dias@ku.ac.ae }

\corres{Correspondence: sandro.a.magalhaes@inesctec.pt (S.A.M.); fbsantos@inesctec.pt (F.N.d.S.)}

\abstract{
The development of robotic solutions for agriculture requires advanced perception capabilities that can work reliably in any crop stage. For example, to automatise the tomato harvesting process in greenhouses, the visual perception system needs to detect the tomato in any life cycle stage (flower to the ripe tomato). The state-of-the-art for visual tomato detection focuses mainly on ripe tomato, which has a distinctive colour from the background. This paper contributes with an annotated visual dataset of green and reddish tomatoes. {This kind of dataset is uncommon and not available for research purposes. This will} enable further developments in {edge artificial intelligence for {in situ} and in real-time} visual tomato detection {required for the development of harvesting robots}. Considering this dataset, five deep learning models were selected, trained and benchmarked to detect green and reddish tomatoes grown in greenhouses. Considering our robotic platform specifications, only the Single-Shot MultiBox Detector (SSD) and YOLO architectures were considered. The results proved that the system can detect green and reddish tomatoes, even those occluded by leaves. SSD MobileNet v2 had the best performance when compared against SSD Inception v2, SSD ResNet 50, SSD ResNet 101 and YOLOv4 Tiny, reaching an F1-score of \SI{66.15}{\percent}, an mAP of \SI{51.46}{\percent} and an inference time of \SI{16.44}{\milli\second} with the NVIDIA Turing Architecture platform, an NVIDIA Tesla T4, with \SI{12}{GB}. YOLOv4 Tiny also had impressive results, mainly concerning inferring times of about \SI{5}{\milli\second}.
}

 % List three to ten pertinent keywords specific to the article, yet reasonably common within the subject discipline.

\begin{document}
%%%%%%%%%%%%%%%%%%%%%%%%%%%%%%%%%%%%%%%%%%

%%%%%%%%%%%%%%%%%%%%%%%%%%%%%%%%%%%%%%%%%%
\section{Introduction}

%Protected agriculture is a field of agriculture which allows producers to increase the yield of their cultures in places and seasons where it is not possible~\cite{jensen1995protected}. Despite there are different kinds of protected agriculture (such as mulching, high tunnels, hydroponics, between others)~\cite{jensen1995protected}, greenhouses~\cite{hanan1997greenhouses} are the most common type. Greenhouses are fixed structures covered with glass or plastic~\cite{jensen1995protected,hanan1997greenhouses}. For yielding these features, they can have climate control, water irrigation control, as well as, artificial illumination~\cite{Shi2019}. Therefore, this kind of agricultural production, when well applied, allow yield increasing or producing non-common plants and vegetables during the whole year.

%Although protected agriculture allows continuous production during the year, the labour need is not constant. Plants have a self-sufficient growing step, i.e., they do not need human labour to support their growing process. So, human labour is only needed to track cultures health and to ensure that greenhouses and cultures have the best conditions. Therefore, human labour needs are seasonal, increasing during the harvesting, pruning, planting and seeding seasons. One of the disadvantages of seasonal labouring needs is the creation of precarious works, which sometimes means the human labour shortage. 

Tomatoes, grown in different crop systems, are the world's second-most harvested vegetable and the leader among greenhouse vegetables~\cite{FAOSTAT2020}. In the last few decades, greenhouse tomato cultivation in several systems has increased worldwide because it has the advantage of enabling high productivity and stable supply throughout the year. 

While the value of greenhouse tomatoes is high on a per-unit basis, the costs are also high, mainly due to the labour costs. Greenhouse's manual operations account for up to 50\% of the total greenhouse production costs, a large part of these costs being absorbed by the manual tomato harvesting, which requires \add{{\SIrange{700}{1400}{\hour\per\year\per\hectare}} according to the cropping system~{\cite{SousaFerreira2017,Camara-Zapata2019}}}. Hence, manual harvesting of tomatoes is, actually, a challenge due to the global labour shortage and precarious working conditions~\cite{Camara-Zapata2019,Fongmul2020,Mitaritonna2020}. Moreover, farmers need to secure additional workers during harvest seasons because the manpower requirements are higher than usual during this time~\cite{Mitaritonna2020}. Therefore, reducing the amount of labour in a greenhouse company is an important topic at the moment, and many companies are looking for automated solutions, such as harvesting robots~\cite{Mitaritonna2020}.

\begin{figure}[H]
 \centering
 \hfill
 \begin{subfigure}[b]{0.24\textwidth}
  \centering
  \includegraphics[width=\textwidth]{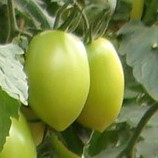}
  \caption{Green tomato}
  \label{fig:green-tomato}
 \end{subfigure}
 \hfill
 \begin{subfigure}[b]{0.24\textwidth}
  \centering
  \includegraphics[width=\textwidth]{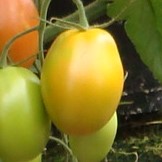}
  \caption{Reddish tomato}
  \label{fig:good-tomato}
 \end{subfigure}
 \hfill
 \begin{subfigure}[b]{0.24\textwidth}
  \centering
  \includegraphics[width=\textwidth]{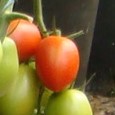}
  \caption{Red tomato}
  \label{fig:ripe-tomato}
 \end{subfigure}
 \vspace{6pt}
 
 \caption{Tomatoes' ripeness levels: (\textbf{a}) physiological or horticultural maturation; (\textbf{b}) early phase of ripening; and (\textbf{c}) ripened tomato.}
 \label{fig:ripeness}
 \hfill
\end{figure}

The development of harvesting robots to work in greenhouses is always challenging because robots have to work in unstructured environments and perform uncertain tasks. In the case of harvesting tomato, the sensing mechanism has to detect fruit in the presence of various disturbances in an unpredicted heterogeneous environment included different arrangements of plant sizes and shapes, stems, branches, leaves, variable colour, under leaves, sun glare and variable light conditions~\cite{Bac2014,Kapach2012}. Even more, as a climacteric fruit, according to the technological objectives, the tomato can be harvested at the physiological maturity phase (green colour), completing the ripening after harvest, or it can be harvested later at different stages of ripening in which it already has a reddish colour~\cite{Giovannoni2001}. When to harvest will depend on how the tomato will be handled and used. Fresh market fruit for local consumers can be picked red, while fruit that will be transported long distances should be harvested at early maturation with green colour (Figure~\ref{fig:ripeness}).

Accurately identifying and detecting the mature fruit or fruit bunches comprise a key technique of harvesting robot research, which has recently received considerable scientific scrutiny. The performance of tomato harvesting robots has been greatly improved by the use of Artificial Intelligence (AI) tools mainly in tomato detection on images acquired in varying environmental and growth conditions such as fruit partially hidden by leaf or stem, state of ripeness (coloration) and light conditions.

The current State-of-the-Art (SoA) explores and proposes different strategies to classify and detect tomatoes based on RGB images. However, in a general overview, these strategies are mostly framed in Machine Learning (ML) applications. These strategies are divided essentially based on the application of the most classical ML algorithms, using, for instance, Support Vector Machines (SVMs)~\cite{Liu2019} or Deep Learning (DL) strategies~\cite{Zhang2018}. DL is based on the training of Artificial Neural Network (ANN) models to recognise features on annotated images (usually referred to as the training set). Then, the trained model is used to detect the trained objects on new images. The state-of-the-art identifies different ANN model structures, such as: 
\begin{enumerate}[label=(\roman*)]
 \item Convolutional Neural Networks (CNNs);
 \item Single-Shot MultiBox Detectors (SSDs).
\end{enumerate}

In comparison to other state-of-the-art ANN structures, the SSD~\cite{Liu2016} aims to be faster. Besides, the SSD does not compromise on the detection accuracy~\cite{Liu2016}. 

This work purposes the development of a vision system to recognise tomatoes in real scenarios by using a dataset of tomatoes grown in a greenhouse. The system explores DL models to be run in a TensorFlow Processing Unit (TPU)~\cite{Jouppi2017}, to assure a high-speed tomato detection system. Currently, only the SSD and some YOLO models can run inside the TPU, and we benchmarked five SSD and YOLO pre-trained models in the COCO dataset~\cite{Lin2014} and the Open Images Dataset (OID)~\cite{Kuznetsova2020}: 
\begin{enumerate}[label=(\roman*)]
 \item SSD MobileNet v2; 
 \item SSD Inception v2; 
 \item SSD ResNet 50; 
 \item SSD ResNet 101;
 \item YOLOv4 Tiny~\cite{Huang_2017_CVPR}.
\end{enumerate}

The paper's structure is based on five sections, complemented by the current Introduction section. Section~\ref{sec:state-of-art} reviews of the related work that contributes to the experiment and states the needed background to understand it. Section~\ref{sec:material-methods} explains the experiment, i.e., how we gathered the data and pre-processed them and how the experiment was performed. Section~\ref{sec:results} presents the results and performs a deep analysis to understand the best deep learning models for tomatoes' detection. Finally, Section~\ref{sec:conclusion} summarises the experiments and the results and indicates the future work to improve these results towards the online detection of tomatoes for harvesting or monitoring purposes.

\section{State-of-the-art}
\label{sec:state-of-art}

\subsection{Literature Review}

This section presents various algorithms, methods and techniques that were proposed and used by different authors regarding fruit detection, more specifically the tomato (Table~\ref{tab:LR}).

In recent years, machine learning, and especially deep learning, techniques in fruit detection has been increasingly used and tested~\cite{Yin2009,Huang2012,Arefi2011,Zhang2015,Benavides2020,Malik2018,Zhu2012,Xiang2013,Yamamoto2014,Zhao2016,Liu2019,Wu2019,Lili2017,Xu2020,Liu2020,Sun2020,Mu2020,AGRIVITA2499,Yuan2020}. Unlike conventional methods, machine learning is a more robust and accurate alternative with a better response to problems such as occlusion and green tomato detection. The problem of green tomato detection is rarely studied due to the difficulty of segmentation and differentiating it from the background, as they have similar colours. This can be observed by the comparison made by~\citet{EAlamSiddiquee2020}. They compared a machine learning method, known as the ``cascaded object detector'', with a system that combines more traditional methods of image processing: ``colour transformation'', ``colour segmentation'', and ``circular Hough transformation'', in the detection of ripe tomatoes. The results showed that the accuracy of the machine learning method is 95\% better than conventional methods. 

For the detection and segmentation purposes of tomatoes in plants, authors usually consider the plants' canopy as the Region of Interest (RoI) (see {Figure~\ref{fig:img-split}).} {In this} RoI, besides the fruits, other structures may be seen that make the fruits' detection difficult---relevant to estimate the fruit location. These other structures may occlude or overlap the tomato/fruit, and {this creates some challenges for the algorithms or processes that are responsible for the fruit detection and segmentation}. These {challenges become greater} in the early ripening stages, due to the high colour correlation {between leaves and tomatoes}. {Despite this fact, the most common and relevant research found in the literature considers the harvesting period in the late maturation stage of tomatoes (where the tomato is already red), so the} colour is therefore a feature used recurrently to differentiate the objects to be detected~\cite{Yin2009,Huang2012,Zhao2016, Arefi2011,Feng2015,Zhang2015,Benavides2020,Malik2018}. Considering the case of fruit detection and segmentation, the authors try to distinguish it from everything external and the background, which at the crop level, can be very complex. Several colour spaces such as Hue, Saturation and Intensity (HSI)~\cite{Arefi2011,Zhang2015}, CIELAB (or L*a*b*)~\cite{Yin2009,Huang2012,Zhao2016} and RGB~\cite{Arefi2011,Feng2015,Zhang2015,Benavides2020} are used to extract this feature. Besides, mathematical morphology approaches~\cite{Dougherty1997} combined with machine learning techniques have also been used in fruit detection in occlusion and overlap situations~\cite{Yin2009,Huang2012,Arefi2011,Zhang2015,Benavides2020,Malik2018,Zhu2012,Xiang2013,Yamamoto2014,Zhao2016,Liu2019,Wu2019,Lili2017}.

For developing a harvesting robot in greenhouses, \citet{Yin2009} segmented ripe tomatoes through K-means clustering using the CIELAB colour space, recording an average task execution time of 10.14~\si{\second}. \citet{Huang2012} used the CIELAB colour space to segment and localise ripe tomatoes in a greenhouse and bi-level partition fuzzy logic entropy to discriminate the fruits from the background. They did not evaluate the performance of the algorithm. \citet{Zhao2016} developed a detection algorithm capable of recognising green, or intermediate, and ripe tomatoes. First, the images of component a* and the images of component L* were extracted from the colour space L*a*b* and the luminance of the Quadrature-phase (YIQ) colour space, respectively. Then, wavelet transformation was adopted to merge the images at the pixel level, which combined the information from the two original images. Finally, to differentiate the fruits from the background, an adaptive threshold algorithm was used to obtain the optimal threshold. The evaluation tests proved a detection rate of 93\% of tomatoes. 

\citet{Arefi2011} proposed an algorithm for the recognition of ripe tomatoes through a combination of the RGB, HSI and YIQ colour spaces and the morphological characteristics of the image. The algorithm obtained a total accuracy of 96.36\% when tested in a greenhouse with artificial lighting. \citet{Feng2015} developed a ripe tomato harvesting robot for a greenhouse, whose identification and location of fruits consisted of the transformation of the RGB colour space images into an HSI colour model to identify and locate the fruits. The robot performed this task in 4~\si{\second}, and the harvest process had a success rate of 83.9\%. \mbox{\citet{Zhang2015},} aiming to detect ripe tomatoes, also converted the RGB colour space into an HSI colour space. The ripe tomato region was cut based on the grey distribution of the H component using the threshold method. The Canny operator~\cite{Canny1986} was used to detect the edges. After a corrosive expansion, the coordinates of the centre of the tomato were marked. The results were not quantified. \citet{Benavides2020} designed a computer vision system for the detection of ripe tomatoes in greenhouses. The segmentation of the fruit was mainly done based on the colour and edges of the fruit, using the R component of the RGB images and the Sobel operator~\cite{Gupta2013}, respectively. Clustered tomatoes were detected with a precision of 87.5\% and beef tomatoes with 80.8\%. \citet{Malik2018} presented a ripe tomato detection algorithm based on the HSV (Hue, Saturation, Value) colour space and the watershed segmentation method. For removing the background and detecting only ripe tomatoes, the HSV colour space was used, and through morphological operations, it was possible to modify the detected fruits. The watershed segmentation algorithm was implemented to ``separate'' the clustered fruits. The combination of these two methods led to a precision of 81.6\%. 

\citet{Zhu2012} combined the mathematical morphology with a Fuzzy C-Means (FCM)-based method for the detection of ripe tomatoes in a greenhouse, with no results reported. Based on the mathematical morphology, \citet{Xiang2013} tested a ripe cluster tomato recognition algorithm. The algorithm was divided into four fundamental steps: tomato image segmentation, performed based on a normalised colour difference; recognition of the clustered region according to the length of the longest edge of the minimum enclosing rectangle of the tomato region; clustered regions, in a binary image, were processed by an iterative erosion course to separate each tomato in this clustered region, and every seed region in the clustered region acquired by the iterative erosion was restored using a circulatory dilation operation. At a distance of \SI{500}{\milli\metre}, they achieved a detection rate of 87.5\%, while from \SIrange{300}{700}{\milli\metre}, the rate dropped to 58.4\%. 

\citet{Yamamoto2014} used different machine learning techniques to detect and distinguish the different stages of tomato ripeness. The proposed method consisted of three steps: pixel-based segmentation conducted to roughly segment the pixels of the images into classes composed of fruits, leaves, stems, and background; blob-based segmentation to eliminate the wrong classifications generated in the first step; and finally, X-means clustering was applied to detect fruits individually in a fruit cluster. The results indicated a precision of 88\%. \citet{Zhao2016a}, to detect ripe tomatoes, extracted the Haar-like features of the grey-scale image, classifying them with the AdaBoost classifier. The false negatives that derived from this classification were eliminated using a colour analysis approach based on the average pixel value (APV). The results showed that the combination of AdaBoost classification with the colour analysis allowed a 96\% detection rate, although 10\% were false negatives and 3.5\% of the fruits were not detected. \citet{Liu2019} proposed an algorithm for the detection of greenhouse ripe tomatoes, where the Histogram of Oriented Gradients (HOG) descriptor was used to train a Support Vector Machine (SVM) classifier. A coarse-to-fine scanning method was developed to detect the fruit, followed by the proposed False Colour Removal (FCR) method to eliminate false-positive detections. The Non-Maximum Suppression (NMS) method was finally used in order to merge the overlapping results. The algorithm was able to detect the fruits with an accuracy of 94.41\%. \citet{Wu2019} developed a greenhouse ripe tomato detection algorithm for a harvesting robot, through a method that combines the analysis and selection of multiple features, a Relevance Vector Machine (RVM) classifier and a bi-layer classification strategy. The algorithm demonstrated an accuracy of 94.90\%. \citet{Lili2017}, developing a greenhouse harvest robot for tomatoes, used the Otsu segmentation algorithm~\cite{Otsu1979} for the automatic detection of ripe tomatoes, obtaining success rates of 99.3\%. 

In the most recent SoA, the interest in DL strategies has been growing~\cite{Xu2020,Liu2020,Sun2020,Mu2020,AGRIVITA2499,Yuan2020}. This interest is due to the higher computability rate of the most recent computers and new edge computing devices dedicated to running DL models, as the TPU. Among the different DL architectures, the You Only Look Once (YOLO) models~\cite{yolo,yolov3} seem to be the most common ones~\cite{Xu2020,Liu2020}. However, Convolutional Neural Network (CNN) structures also have their place due to their high accuracy, besides the long inference time~\cite{Liu2016}. This issue allows the appearance and use of the SSD adaptation of these models~\cite{Liu2016}. Among these adaptations, we can find the insertion of new layers (to increase the network resolution) and pruning of the output layers to fit the network classes, change the detection or feature extraction layers. 

\citet{Xu2020} improved the YOLOv3-tiny method to obtain a faster and more accurate detection of ripe tomatoes. The accuracy of the model was increased by improving the backbone network, and the image enhancement allowed better detection in more complex scenarios. The results obtained showed that the F1-score of the proposed model was 91.92\%, which was 12\% higher than the unmodified YOLOv3-tiny method. \citet{Liu2020} used the YOLOv3 detection model to create the YOLOTomato model, which was possible to achieve due to the incorporation of the dense architecture for feature extraction and the replacement of the traditional R-box by the proposed C-box. In scenarios with moderate occlusions, the model obtained a detection rate of 94.58\%, 4\% more than in scenarios with severe occlusions. In order to overcome overlaps and occlusions, \citet{Sun2020} developed a detection method based on CNN, more specifically the feature pyramid network method. By comparing this method with traditional Faster Region-based Convolutional Neural Network (R-CNN) models, the proposed method improved the detection rate from 90.7\% to 99.5\%. \citet{Mu2020} built a tomato detection model capable of detecting green tomatoes in greenhouses, regardless of possible occlusions. The model uses a pre-trained Faster R-CNN structure with the deep convolutional neural network ResNet-101 based on the Common Objects in Context (COCO) dataset, which was then fine-tuned for tomato detection, reaching an accuracy of 87.83\%. 

As will be studied in this paper, the Deep Learning Single-Shot Multibox Detector (SSD) model promises a substantial improvement in fruit detection. Therefore, it has been increasingly studied, since it can capture the information of an object and its anti-interference, as well as directly complete the localisation and the classification task in just one step. 

This improvement was demonstrated by \citet{AGRIVITA2499} who designed a computer visualisation system to evaluate the growth of tomato plants through the detection of fruits and flowers. Two deep learning models were used: R-CNN and SSD. The fruit detection accuracy of the R-CNN model was only 19.48\%, while the SSD model showed a much higher detection rate of 95.99\%. To detect cherry tomatoes in greenhouses, whether ripe, green or intermediate, \citet{Yuan2020} developed an SSD-based algorithm. After building the datasets, they were used to train and develop network models. For studying the effect of the base network, one of the experiments was tested on different networks, such as VGG16, MobileNet and Inception V2. The results indicated that the Inception V2 network obtained the best performance with an accuracy of 98.85\%.

% start a new page without indent 4.6cm
\clearpage
\end{paracol}
\nointerlineskip

\begin{figure}[H]
 \widefigure
 \includegraphics[width=0.95\textwidth]{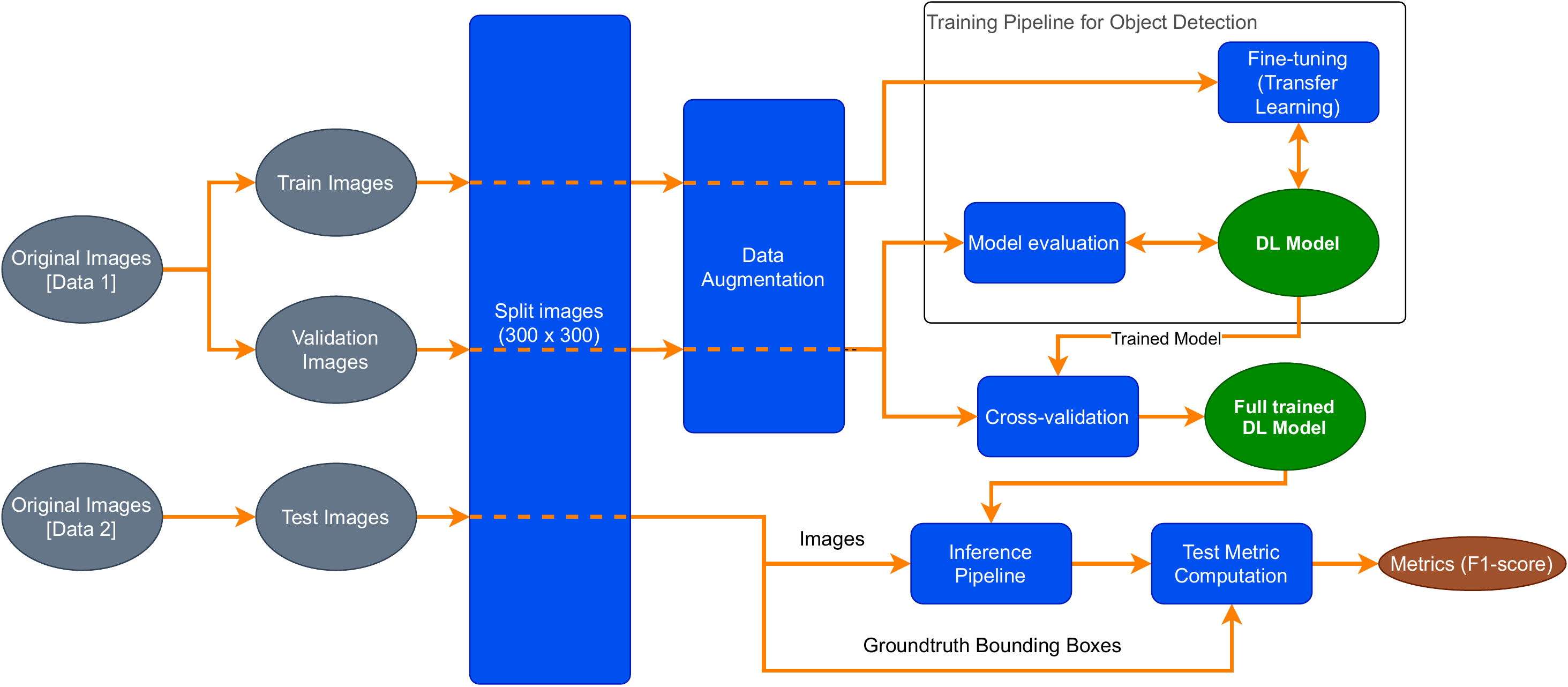}
 \caption{Overview of the performed methods. Training and evaluation pipeline.}
 \label{fig:dl_overview}
\end{figure}

\begin{specialtable}[H] 
\small
\widetable
\caption{Algorithms, methods and techniques proposed by different authors regarding tomato detection at different ripeness levels (N/A---Not Available).} \label{tab:LR} 
\begin{tabular}{l@{\hspace{0.2cm}}c@{\hspace{0.1cm}}c@{\hspace{0.01cm}}c@{\hspace{0.1cm}}c@{\hspace{0.2cm}}}
 \toprule
 \textbf{Method} & \textbf{Tomato Ripeness} & \textbf{Accuracy} & \textbf{Inference Time} & \textbf{Authors/Year}
 \\ \midrule 
 
L*a*b* colour space and& Ripe & N/A & \SI{10.10}{\second} & {\citet{Yin2009} \citeyear{Yin2009}}\\
 K-means clustering & &&&\\
 L*a*b colour space and bi-level\\ partition fuzzy logic entropy & Ripe & N/A & N/A & {\citet{Huang2012} \citeyear{Huang2012}}\\
 L*a*b colour space and & Green, intermediate & \SI{93}{\percent} & N/A & \citet{Zhao2016} \citeyear{Zhao2016}\\
Threshold algorithm & and ripe & &&\\
 RGB, HSI and YIQ colour spaces\\ and morphological characteristics & Ripe & \SI{96.36}{\percent} & N/A & {\citet{Arefi2011} \citeyear{Arefi2011}}\\
 RGB colour space images into an HSI & Ripe & \SI{83.9}{\percent} & \SI{4}{\second} & {\citet{Feng2015} \citeyear{Feng2015}}\\
 colour model & Ripe & \SI{83.9}{\percent} & \SI{4}{\second} & {\citet{Feng2015} \citeyear{Feng2015}}\\
 RGB colour space into an HSI \\ colour space, threshold method \\and Canny operator & Ripe & N/A & N/A & {\citet{Zhang2015} \citeyear{Zhang2015}}\\
 R component of the RGB images\\ and Sobel operator & Ripe & Clustered tomatoes: \SI{87.5}{\%} & N/A & {\citet{Benavides2020} \citeyear{Benavides2020}}\\
&& Beef tomatoes: \SI{80.8}{\%} & & \\
 HSV colour space and watershed & Ripe & \SI{81.6}{\percent} & N/A & {\citet{Malik2018} \citeyear{Malik2018}}\\
 segmentation method & &&&\\
 Mathematical morphology and\\ Fuzzy C-Means-based method & Ripe & N/A & N/A & {\citet{Zhu2012} \citeyear{Zhu2012}}\\
 Mathematical morphology, \\ difference and iterative erosion course & Ripe & \SI{50}{\centi\metre}--\SI{87.5}{\percent} 30 to & N/A & {\citet{Xiang2013} \citeyear{Xiang2013}}\\
 Normalised colour&& 70~\si{\centi\metre}--\SI{58.4}{\percent} && \\
 Pixel-based segmentation, blob-based\\ segmentation and X-means clustering & Green, intermediate& \SI{88}{\percent} & N/A & {\citet{Yamamoto2014} \citeyear{Yamamoto2014}}\\
 & and ripe & &&\\
 Haar-like features of grey-scale image\\ and AdaBoost classifier & Ripe & \SI{96}{\percent} & N/A & {\citet{Zhao2016a} \citeyear{Zhao2016a}}\\
 Histograms of oriented gradients & Ripe & \SI{94.41}{\percent} & N/A & {\citet{Liu2019} \citeyear{Liu2019}}\\
 and SVM & Ripe & \SI{94.41}{\percent} & N/A & {\citet{Liu2019} \citeyear{Liu2019}}\\
 Analysis and selection of \\multiple features, & Ripe & \SI{94.90}{\percent} & N/A & {\citet{Wu2019} \citeyear{Wu2019}}\\
 RVM and bi-layer classification\\ strategy &&&&\\ 
 Otsu segmentation algorithm & Ripe & \SI{99.3}{\percent} & N/A & {\citet{Lili2017} \citeyear{Lili2017}} \\
 Improved YOLOv3-tiny method & Ripe & F$_1 =$ \SI{91.92}{\percent} & N/A & {\citet{Xu2020} \citeyear{Xu2020}}\\
 YOLOv3 detection model to create the& Green, intermediate & \SI{94.58}{\percent} & N/A & {\citet{Liu2020} \citeyear{Liu2020}}\\
 proposed YOLOTomato model &and Ripe &&&\\
 Feature pyramid network & Green, intermediate & \SI{99.5}{\percent} & N/A & {\citet{Sun2020} \citeyear{Sun2020}}\\
 & and ripe & &&\\
 Faster R-CNN structure with the & Green & \SI{87.83}{\percent} & N/A & {\citet{Mu2020} \citeyear{Mu2020}} \\
  deep CNN ResNet-101 &&&&\\
 Comparison: R-CNN vs. SSD & Green, intermediate & {R-CNN: \SI{19.48}{\percent}} & N/A & {\citet{AGRIVITA2499} \citeyear{AGRIVITA2499}} \\
 & and Ripe & SSD: \SI{95.99}{\percent} &&\\
 SSD-based algorithm used to train & Green, intermediate & Best performance is & N/A & {\citet{Yuan2020} \citeyear{Yuan2020}}\\
and develop network models & and ripe &Inception V2 (\SI{98.85}{\percent}) & & \\ 
 such as VGG16, MobileNet,\\ Inception V2 & &&&\\ 
\bottomrule
\end{tabular}
\end{specialtable}
\begin{paracol}{2}
%\linenumbers
\switchcolumn

\vspace{-6pt}

\subsection{Background}

\subsubsection{SSD Architecture}

The Single-Shot MultiBox Detector (SSD) belongs to the One-Step Framework, also known as the Regression or Classification Based Framework, just like YOLO or RetinaNet~\cite{zhao2019object,lin2017focal}. With such frameworks, there is an explicit mapping between pixel values, bounding box coordinates and class probabilities, unlike Region Proposal-Based Frameworks, e.g., Faster RCNN. Therefore, compared to Faster RCNN and the same category of architectures, the SSD has lower inference times to the point of achieving real-time performance. %(Figure~\ref{DL_comp}) ~\cite{Liu2016,Huang_2017_CVPR}.

%\begin{figure}[H]
%\centering
%\includegraphics[scale=0.5]{figures/SSD_explain/DL_models_comparison.png}
%\caption{Comparison between several object detection architectures and %backbones~\cite{Huang_2017_CVPR}}
%\label{DL_comp}
%\end{figure}

The SSD architecture, depicted in Figure~\ref{SSD_arch}, is composed of two main parts: feature extraction and object detection. For the first one, a state-of-the-art classification model is usually used (e.g., the VGG16~\cite{DBLP:journals/corr/SimonyanZ14a} network as in Figure \ref{SSD_arch}), but others like ResNet~\cite{He_2016_CVPR} or MobileNet~\cite{Sandler_2018_CVPR} are also possible. The feature extractor is called the backbone, and its purpose is to generate high-level feature maps from the input image. Besides the backbone, the SSD adds six extra feature maps with a decreasing spatial dimension; see Figure~\ref{SSD_arch}~\cite{Liu2016,Huang_2017_CVPR}.

\begin{figure}[H]
%\centering
\includegraphics[width=\linewidth]{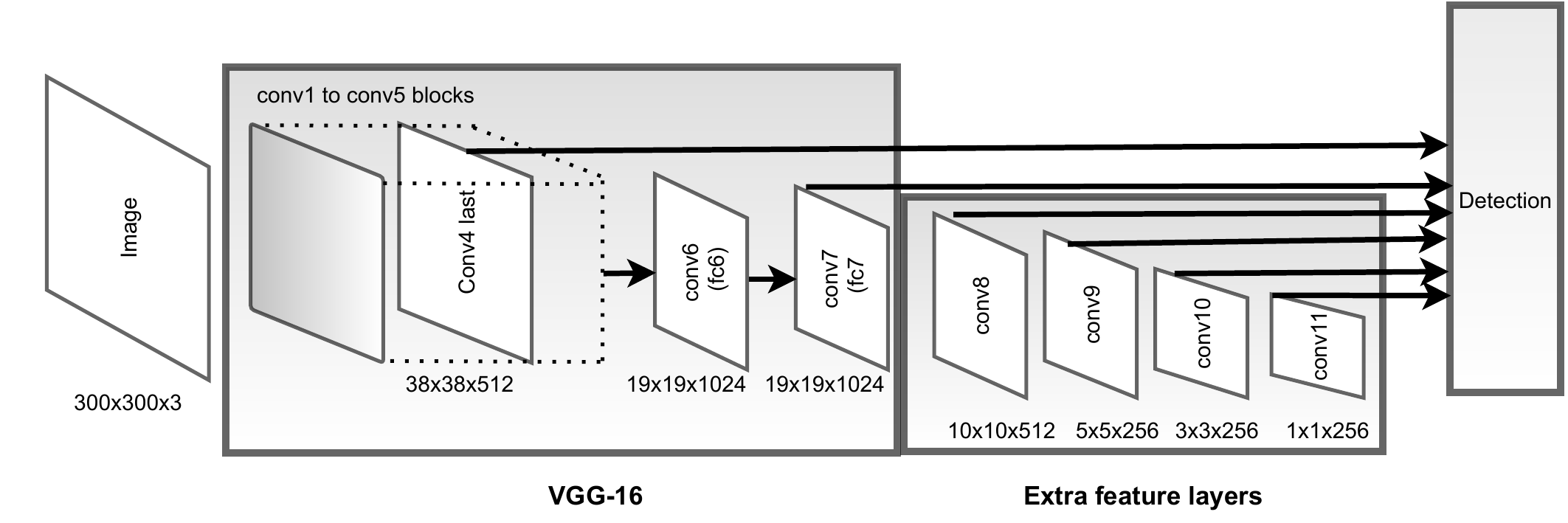}
\caption{Scheme for the SSD architecture using VGG16 as the backbone.~{Adapted with permission from ref.\cite{Liu2016}.} }%S.A.M: The figures were rebuilt and adapted accordingly with the licenses
\label{SSD_arch}
\end{figure}

For the second part, the SSD relies on a set of default anchors (i.e., bounding boxes) (\mbox{Figure~\ref{BB_anchor}),} with different aspect ratios and scales, thus reducing the possible amount of shapes that bounding boxes can assume~\cite{zhao2019object}. A convolution layer is responsible for predicting, for each convolution operation and for each default anchor box, the location offset for that anchor and confidence scores for each class on the dataset. This convolution layer is applied to the extra feature maps. In the case of the VGG16 backbone, this layer is also applied to the \verb|Conv4_3| output~\cite{Liu2016}. Fusing the predictions made from the feature maps, each with a different resolution, allows detecting objects of different sizes. From Figure~\ref{SSD_arch}, using the feature maps towards the right will result in detecting larger objects, and vice versa~\cite{zhao2019object}.

\begin{figure}[H]
%\centering
\includegraphics[scale=0.6]{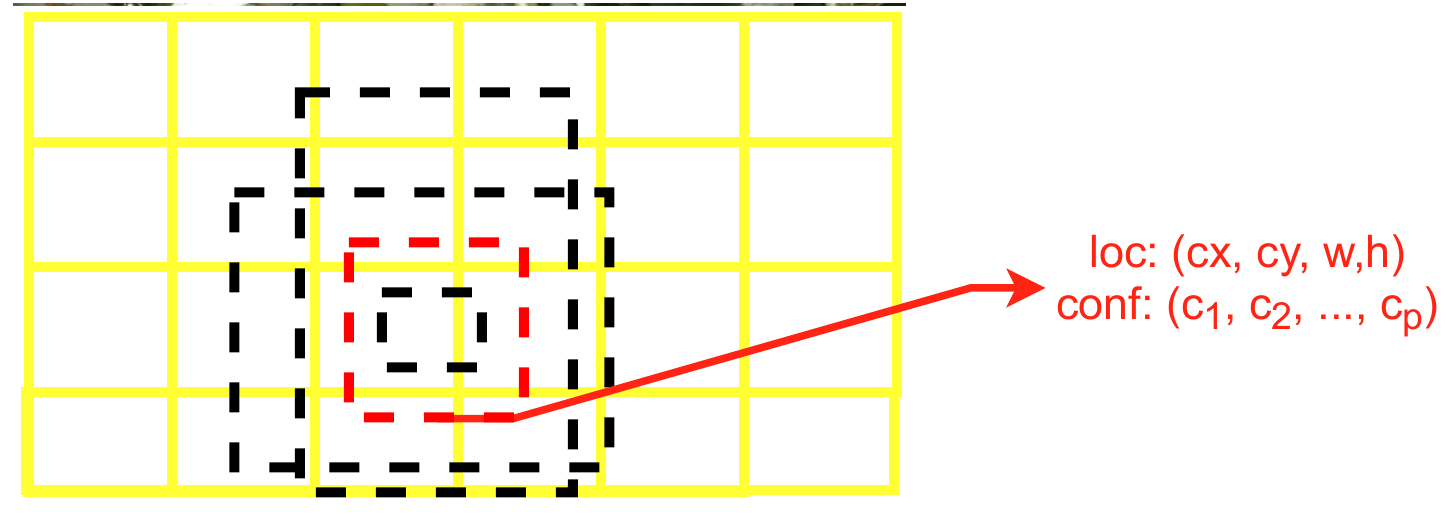}
\caption{Anchor box shapes used in the SSD architecture.~{Adapted with permission from ref.\cite{Liu2016}.} }
\label{BB_anchor}
\end{figure}
%S.A.M: The figures were rebuilt and adapted accordingly with the licenses
In the end, many detections will be predicted, so Non-Maximum Suppression (NMS) is applied to keep the highest rated bounding boxes. As concerns training, a weighted sum between localization loss (e.g., smooth L1) and confidence loss (e.g., softmax) is used~\cite{zhao2019object}.

In order to improve SSD performance, the following measures can be used: choose default anchors with scales and aspect ratios according to the problem under analysis; perform data augmentation; and use hard negative mining. Even though, the SSD architecture performs worst in small object detection as these do not appear in all feature maps. This problem can be mitigated by using better feature extractor backbones (e.g., ResNet) and higher resolution input images~\cite{zhao2019object}.

%%%%%%%%%%%%%%%%%%%%%%%%%%%%%%%%%%%%%%%%%%
\section{Materials and Methods}
\label{sec:material-methods}

Figure~\ref{fig:dl_overview} reports an overview of the training and evaluation pipeline to reach a trained DL model. This architecture will be detailed in the following subsections. 

% start a new page without indent 4.6cm
%\clearpage
\end{paracol}
\nointerlineskip
\begin{paracol}{2}
%\linenumbers
\switchcolumn
\vspace{-6pt}

\subsection{Data Acquisition}

Effective harvesting robots need to detect the harvesting target efficiently. In the current study, the robot intends to harvest tomatoes in a greenhouse tomatoes' culture. Commonly used datasets, such as the COCO dataset~\cite{Lin2014}, Open Images Dataset (OID)~\cite{Kuznetsova2020} and KITTI dataset~\cite{Geiger2012}, provide a large amount of data, but only OID has tomatoes in its data. However, these data are not representative of the kind of data class we intended to detect. 

To overcome these issues, a new image dataset of tomatoes was collected from a greenhouse at Barroselas in Viana do Castelo, Portugal. Although all the greenhouses on the campus are not equal, they have a similar configuration: six hedges of tomato plants with 0.9~\si{\metre} between-row spacing and 1.10~\si{\metre} of height (Figure~\ref{fig:greenhouse}), where the robot can travel. Tomatoes detached from the plant that have fallen to the ground are too ripe and should not be harvested.

\begin{figure}[H]
 % \centering
 \includegraphics[width=0.65\textwidth]{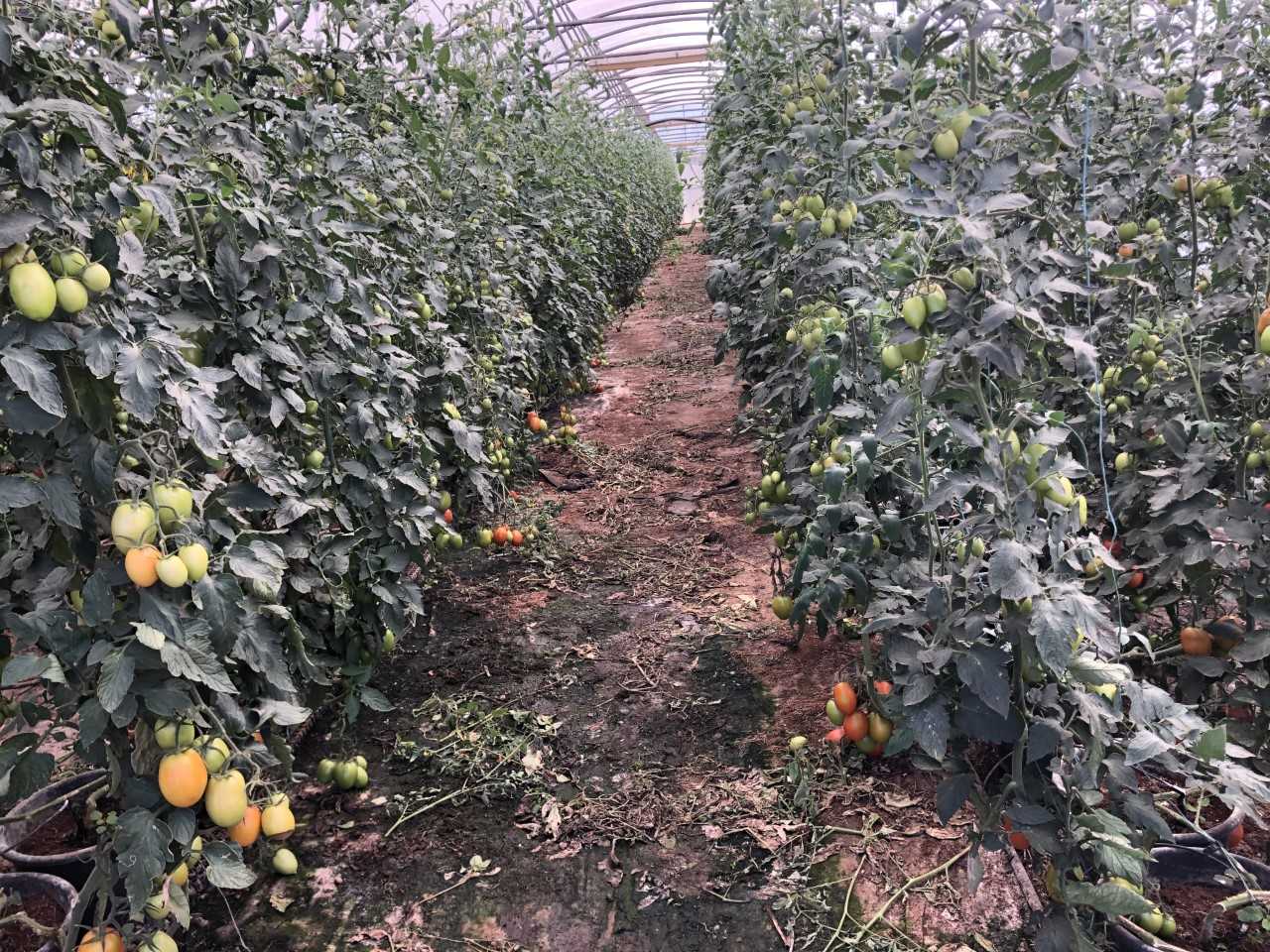}
 \caption{Greenhouses' entrance.}
 \label{fig:greenhouse}
\end{figure}

The mobile robot AgRob v16 (Figure~\ref{fig:robot}) was used for recording images inside the greenhouses to increase the representativeness of the data. {This robot is equipped with a set of sensors commonly used in robotic operations. Therefore, we obtained data acquired in the same conditions as a robot in a regular harvesting operation.} At this stage, a human operator controlled, slowly, the robot through the different halls of the greenhouses, while the robot recorded, into a single ROSBag file, the information provided by its different sensors (cameras, IMU, LiDAR, among others). For the purposes of this study, only RGB images were relevant. The robot was moved along the crop row, keeping a distance between the robot and tomato {plants} from \SIrange{0.4}{0.6}{\metre}. 

\begin{figure}[H]
 %\centering
 \includegraphics[width=0.7\textwidth]{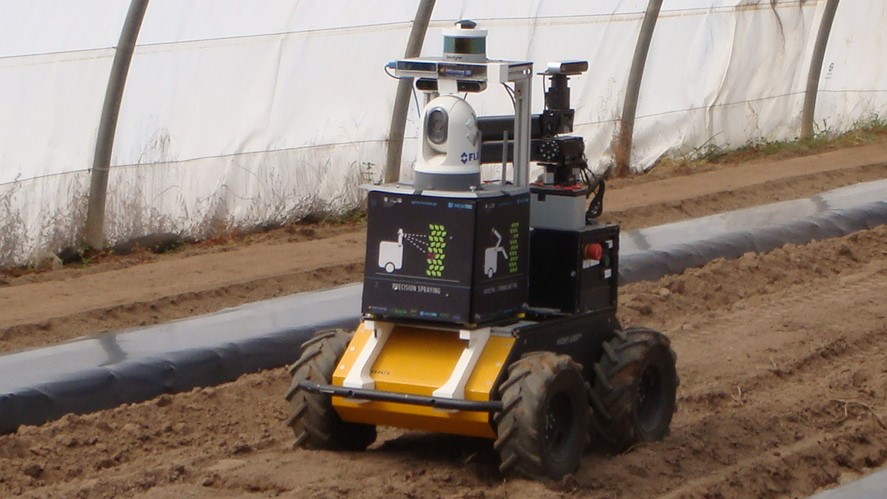}
 \caption{AgRob v16 inside an uncultivated greenhouse.}
 \label{fig:robot}
\end{figure}

As shown in Figure~\ref{fig:robot}, the robot had two stereo cameras. The front camera was mainly used to localise the robot along the hall. However, for harvesting tomatoes, we intended to use a hand-eye strategy~\cite{Hong2005}, which allowed a continuous refinement of the position of the robotic arm with respect to the tomatoes, using active perception~\cite{Bajcsy2017} or gaze control mechanisms~\cite{Ballard1991}. Therefore, the robot used the second stereo camera (ZED; {see} \url{https://www.stereolabs.com/zed/}{, accessed on 25 November 2020}) mounted on an anthropomorphic manipulator at the backside of the robot. The manipulator remained in the rest position as in Figure~\ref{fig:robot}, i.e., looking sideward, towards the tomato plants, during the whole acquisition process. A Jetson Nano ({see} \url{https://developer.nvidia.com/embedded/jetson-nano-developer-kit}{, accessed on 25 November 2020}) Graphics Processing Unit (GPU) connected to the ZED camera in the robotic arm managed the camera and the collected images. After, the GPU sent the images to the onboard computer of the robot, to merge them with the remaining data, collected by the robot.
In summary, a mobile robot collected images of the wall of tomato plants and recorded them as a video in an ROSBag file.
%S.A.M: ok. Added the accessed dates, while writing the manuscript

\subsection{Dataset Generation}

Most DL models are known as supervised ML algorithms. This specificity implies that the training mechanisms for DL models need to be fed with an annotated dataset. In the case of object detection, each annotation identifies its class, size and position. Some annotation types, like the Pascal VOC format, also include some additional features of the annotations as whether the target object is difficult to detect, occluded or truncated. For the current dataset, we used the Pascal VOC format, as used in the Pascal VOC Challenge~\cite{Everingham2010}, due to its simplicity, resuming the annotations for each image in a single XML file. In this case, we ignored the additional features {because} the TensorFlow1 Object Detection Pipeline ignores these features.

First of all, we converted the continuous video with the images of tomatoes into individual sequential images. To avoid high repetitiveness between images in the dataset, we acquired a frame every three seconds, assuring an overlapping ratio of around 60\%. This process resulted in a dataset of \num{297} images with a resolution of $1280 \times 720$ px each.

All the images were manually labelled using the annotation tools CVAT~\cite{boris_sekachev_2020_4009388} and LabelImg~\cite{Tzutalin2015}. These tools allow better management of images and annotations, as well as a collaborative annotation. For this dataset, we only considered the class ``tomato'', ignoring their ripeness, because of the low amount of ripened {tomatoes} (Figure~\ref{fig:ripeness}){, i.e., most of the tomatoes in the dataset were reddish or green (Figure~{\ref{fig:ripeness}})}.

We intended to use TPUs to detect tomatoes online inside the tomatoes' greenhouses. However, TPUs are not compatible with all the ANN models. At the current time, they are only compatible with SSD models~\cite{Liu2016} and the tiny versions of You Only Look Once (YOLO)~\cite{yolov3} model. Additionally, these models cannot process full-sized images, rescaling all the images before processing them. For instance, the pre-trained MobileNet v2~\cite{Huang_2017_CVPR} can only process images of $300 \times 300$ px. Due to this, we split the original images into $300 \times 300$ px images, following the scheme represented in Figure~\ref{fig:img-split}, using \verb|pascal_voc_tools| ({see} \url{https://github.com/wang-tf/pascal_voc_tools}{, accessed on 8 September 2021}). This splitting scheme considered the construction of sub-images of $300 \times 300$ px with a minimum overlapping ratio of \SI{20}{\percent} between sequential images. Splitting full-sized images of $1280 \times 720$ px to $300 \times 300$ px increased the dataset and the quality of the considered images for training. Therefore, this procedure increased the accuracy of the developed system. The Dataset now had \num{5365} images.%S.A.M: Updated access date, accordingly

\begin{figure}[H]
 % \centering
 \includegraphics[width=\linewidth]{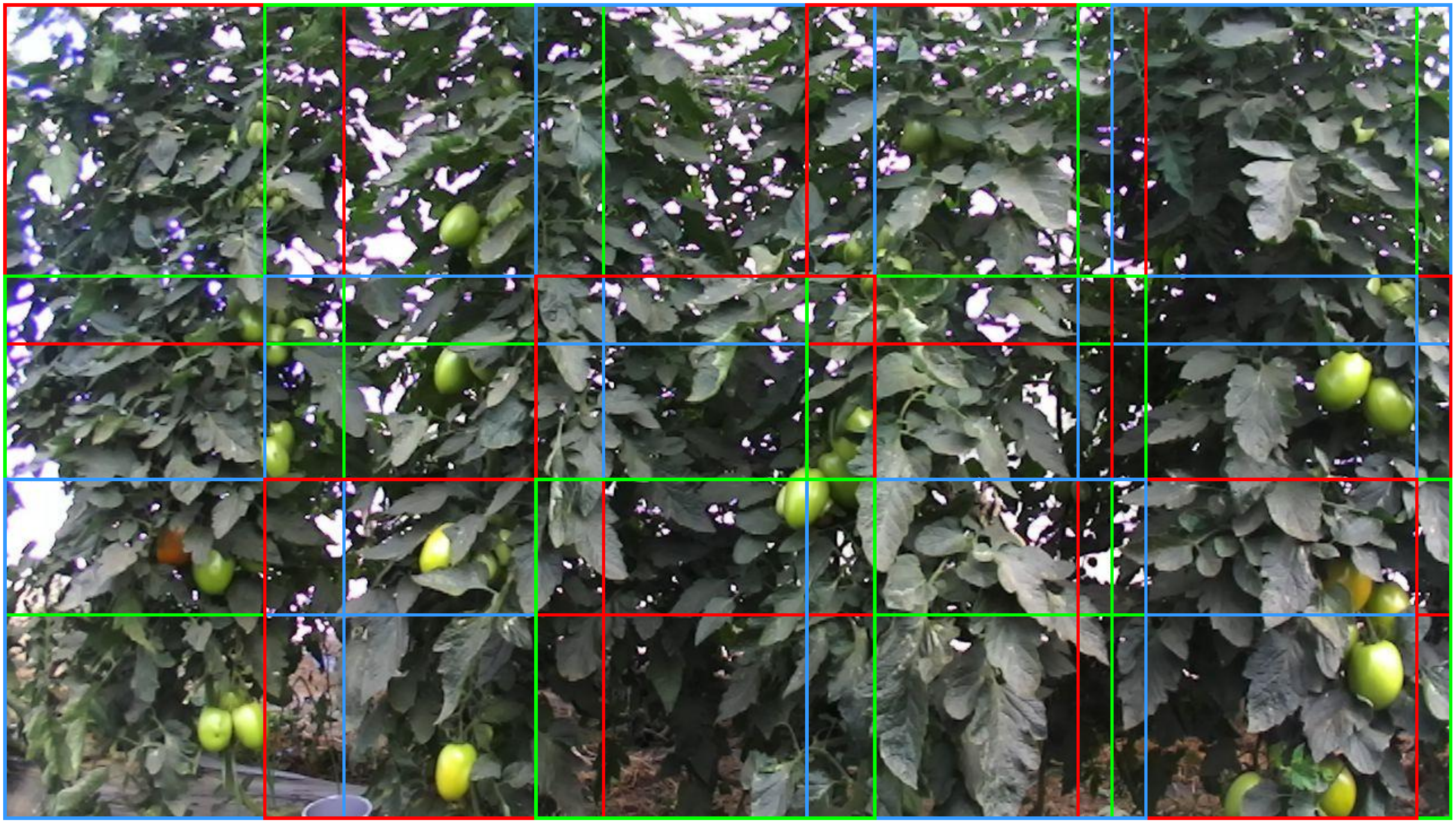}
 \caption{Images split into $300 \times 300$ px images with an overlapping ratio of \SI{20}{\percent}. The different colours are only for reference and distinguishing the different images.}
 \label{fig:img-split}
\end{figure}

Some researches concluded that using augmentation strategies in the original images allowed increasing the dataset size and variability, contributing new information to it~\cite{Shorten2019}. Different kinds of transformations may be applied to an image: 
\begin{enumerate}[label=(\roman*)]
 \item rotation; 
 \item translation; 
 \item scaling; 
 \item hue modification; 
 \item saturation; 
 \item blur; 
 \item noise; 
 \item others, even combinations of transformations. 
\end{enumerate}

Table~\ref{tab:data_augmentation} resumes the different applied transformations to images and their specifications. All transformations were applied with a random factor, to increase the variability of the data. Figure~\ref{fig:augmentation} exemplifies the use of augmentation in images. The augmented dataset resulted in a total of {23,021} images with {61,204} annotations of tomatoes.

For training purposes, we divided the dataset into two sets: training set and validation set. The training set contained {18,417} images with {49,100} annotations. The validation set had {4604} images with {12,104} annotations. 
For the evaluation and testing purposes of the trained models, an external set of annotated images was used. This set was acquired in the same conditions of the training and validation data, but in a different row of the tomatoes' greenhouse. The original set of full-sized images with $1280\times720$~\si{px} had 250 images. For these data, the augmentation was not desired, but we still considered splitting the original images into smaller ones, with $300 \times 300$~\si{px}. This resulted in a set of {2737} images. 
The dataset then had {25,758} images. The acquired data are made publicly available at INESC~TEC {Research Data Repository} ({see} \url{https://rdm.inesctec.pt/} and \url{http://www.doi.org/10.25747/pc1e-nk92}, {updated on 14 May 2021})~\cite{dataset}.%S.A.M: updated RDM name and last updated date

\begin{figure}[H]
 \centering
 \begin{subfigure}[b]{0.24\textwidth}
  \centering
  \includegraphics[width=\textwidth]{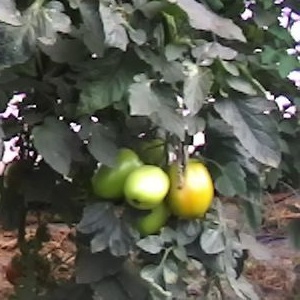}
  \caption{Original}
  \label{fig:aug-original}
 \end{subfigure}%
 \hfill 
 \begin{subfigure}[b]{0.24\textwidth}
  \centering
  \includegraphics[width=\textwidth]{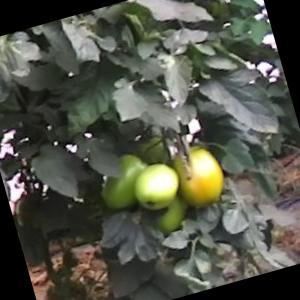}
  \caption{Rotation}
  \label{fig:aug-rotation}
 \end{subfigure}%
 \hfill%
 \begin{subfigure}[b]{0.24\textwidth}
  \centering
  \includegraphics[width=\textwidth]{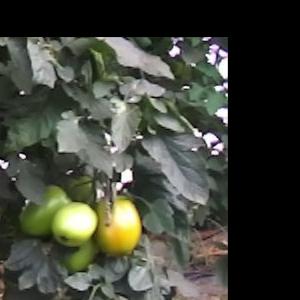}
  \caption{Translation}
  \label{fig:aug-translate}
 \end{subfigure}\\%
 \hfill%
 
 \begin{subfigure}[b]{0.24\textwidth}
  \centering
  \includegraphics[width=\textwidth]{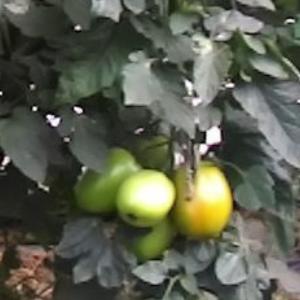}
  \caption{Scaling}
  \label{fig:aug-scale}
 \end{subfigure}%
 \hfill%
\vspace{10pt}
 \begin{subfigure}[b]{0.24\textwidth}
  \centering
  \includegraphics[width=\textwidth]{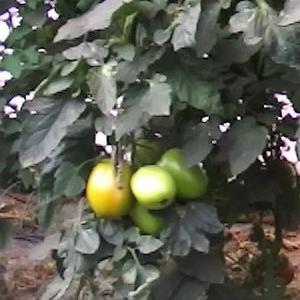}
  \caption{Flip horizontally}
  \label{fig:aug-flip}
 \end{subfigure}%
 \hfill%
 \begin{subfigure}[b]{0.24\textwidth}
  \centering
  \includegraphics[width=\textwidth]{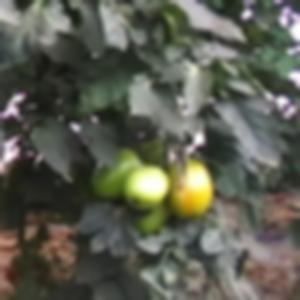}
  \caption{Blurring}
  \label{fig:aug-blue}
 \end{subfigure}
 \hfill%
 \begin{subfigure}[b]{0.24\textwidth}
  \centering
  \includegraphics[width=\textwidth]{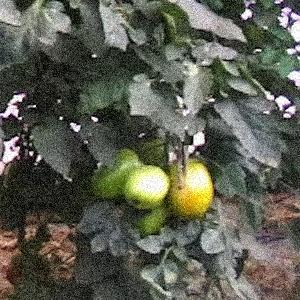}
  \caption{Noising}
  \label{fig:aug-noise}
 \end{subfigure}%
 \hfill%
 \begin{subfigure}[b]{0.24\textwidth}
  \centering
  \includegraphics[width=\textwidth]{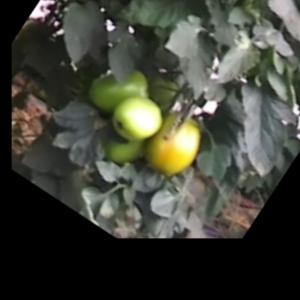}
  \caption{Combination 3}
  \label{fig:aug-combination3}
 \end{subfigure}%
\vspace{10pt} 
 \caption{Example of augmentation applied to an image. (\textbf{h}) {is the random combination of 3 of the other transformations.}}
 \label{fig:augmentation}
\end{figure}
%S.A.M: Thank you. Yes, its a very good description

\begin{specialtable}[H]
 %\centering
 \caption{Transformations applied to the images of the split dataset for data augmentation and the characteristics of those transformations.}
 \label{tab:data_augmentation}
 \begin{tabular}{l@{\hspace{2.1cm}}c@{\hspace{1cm}}}\toprule
 \textbf{Transformation} & \textbf{Value} \\ \midrule
 Rotation & \SIrange{-60}{60}{\degree}\\
 Scaling & \SIrange{50}{150}{\percent} \\
 Translation & \SIrange{0}{30}{\percent} left or right \\
 Flip & Image mirroring\\
 Blur (Gaussian Filter) & $\mathcal{N}(0, 1~\text{to}~3)$ \\
 Gaussian Noise & $\mathcal{N}(0, 0.03 \cdot 255~\text{to}~0.07 \cdot 255)$ \\
 Combination3 &Random combination of three of the\\
 & previous transformations with random values\\
 \bottomrule
 \end{tabular}
\end{specialtable}

\subsection{Training and Evaluating SSD Models}
\label{sec:training}

The state-of-the-art has many frameworks for training and using deep learning algorithms~\cite{Bahrampour2015,Al-Bdour2019,tensorflow2015-whitepaper,darknet13,NEURIPS2019_9015}. However, TensorFlow~\cite{tensorflow2015-whitepaper}, Darknet~\cite{darknet13} and PyTorch~\cite{NEURIPS2019_9015} are the most known and used frameworks. Once we determined that our robot would use a TPU to detect tomatoes in the greenhouse, it needed a trained TensorFlow model.

TensorFlow~\cite{tensorflow2015-whitepaper} is a machine learning system that operates at a large scale and in heterogeneous environments. It is easily scalable and allows researchers and engineers to experiment with and test new ML algorithms. Therefore, TensorFlow supports a large variety of ML algorithms, focusing mainly on DL. TensorFlow is distributed as an open-source framework belonging to Google. It can run in a large variety of applications and devices as a centralised or distributed system. 

During the development of the current evaluation of different models, only TensorFlow 1 had fully compatible tools to train and compile the models to the TPU. Therefore, the training and inference scripts used TensorFlow r1.15.0. Both scripts run in Google Colaboratory (Colab) notebooks ({see} \url{http://colab.research.google.com}) that offer free powerful GPU's and TPU's to train and infer deep learning models. The available GPUs varied each time we initialised a Colab session, but in this case, the NVIDIA Tesla T4 with a VRAM of 12~GB and 7.5 computation capability was the attributed GPU for all sessions.

For benchmarking purposes, we considered four pre-trained SSD models from the TensorFlow database ({see} \url{https://github.com/tensorflow/models/blob/master/research/object_detection/g3doc/tf1_detection_zoo.md}): SSD MobileNet v2, SSD Inception v2, SSD ResNet 50, and SSD ResNet 101 (see Appendix~\ref{ap:a}, Table~\ref{tab:ssd_models_location}, for the location of the different models); and the YOLOv4 Tiny model. The first three models and YOLOv4 Tiny were previously pre-trained using the COCO dataset, and SSD ResNet 101 was pre-trained using OID. For fine-tuning the pre-trained models, we considered the default values of the pre-training pipeline, adjusting the batch size for the capacity of the available GPU, according to Table~\ref{tab:ssd_models_batch}. All training sessions ran for {50,000} epochs, and an evaluation session occurred at every 50 epochs. The experiments with the different models proved that they did not need more than {50,000}~epoch to converge to the best solution in the solution space. In some cases, the models converged after {30,000} epochs. The evaluation session at every 50 epochs followed the standard value used by the pre-trained models. These evaluation sessions allowed monitoring the evolution of the training. If the evaluation loss started to increase while the training loss decreased or remained constant, the deep learning model was over-fit to the training data. This situation did not happen to any trained networks. 

YOLOv4 Tiny is not available for the TensorFlow framework, but is for the Darknet framework. {The YOLOv4 Tiny model learned faster and only needed {{2500}} epochs for the training session}. Darknet had no available validation sessions, so it was not considered.

\begin{specialtable}[H]
 % \centering
 \caption{Training batch size for each model.}
 \label{tab:ssd_models_batch}
 \begin{tabular}{l@{\hspace{4.5cm}}c@{\hspace{4.6cm}}}
 \toprule
 \textbf{SSD Model} & \textbf{Batch Size} \\ \midrule
 SSD MobileNet v2 & 24 \\
 SSD Inception v2 & 32 \\
 SSD ResNet 50 & 8 \\
 SSD ResNet 101 & 8 \\
 YOLOv4 Tiny & 64 \\
 \bottomrule
 \end{tabular}
\end{specialtable}

The literature refers to many different evaluation metrics~\cite{padillaCITE2020,Everingham2010}. During the training process, we used the default pipeline metrics. However, as the COCO dataset and OID use different measures to evaluate the performance of the models, an additional common evaluation metrics pipeline was needed. For better benchmark SSD models, we preferred the metrics used by the Pascal VOC challenge~\cite{Everingham2010} (precision $\times$ recall curve and mean average precision), as implemented by \citet{padillaCITE2020}, and complemented this evaluation with additional metrics: 
\begin{enumerate}[label=(\roman*)]
 \item total recall; 
 \item total precision; 
 \item F1-score.
\end{enumerate}

Recall \eqref{eq:recall} is the ability of the model to detect all the relevant objects, i.e., the ability of the model to detect all the detected bounding boxes of the validation set. Precision \eqref{eq:precision} is the ability of the model to identify only the relevant objects. F1-score \eqref{eq:F1} is the first harmonic mean between recall and precision. True Positives (TPs) are the correct detections of the ground truths. {False Positives (FPs) are the objects that were improperly detected. False Negatives (FNs) are the undetected objects. The number of ground truths can be computed by the sum of the TPs and FNs (TP $+$ FN), and the number of detections is the sum of the TPs and FPs (TP $+$ FP). The detections} are normally validated using the Intersection Over Union (IOU) metric~\cite{padillaCITE2020} considering only the detections with an IOU $\geq t$. For the current benchmark, we considered $t = 50$\%.

\begin{equation}
 \text{Recall} = \dfrac{\text{TP}}{\text{TP} + \text{FN}} \label{eq:recall} 
\end{equation}

\begin{equation}
 \text{Precision} = \dfrac{\text{TP}}{\text{TP} + \text{FP}} \label{eq:precision}
\end{equation} 

\begin{equation}
 \text{F}_1 = 2 \cdot \dfrac{\text{Precision} \times \text{Recall}}{\text{Precision} + \text{Recall}} \label{eq:F1}
\end{equation}

At the end of the training pipeline, all the models still had a free parameter: the confidence rate. The confidence rate is a value that features the certainty in the performed prediction. A prediction with a confidence rate of \SI{50}{\percent} determines that the network is \SI{50}{\percent} sure of the detected or classified object. Robust ANN tends to detect better and with higher confidences, but lower confidences can still have true positives. Therefore, optimising the confidence score is essential to optimise the network performance. The cross-validation technique is a useful technique to optimise this value. In the validation set, we removed all the augmentations and computed the F1-score for all the confidence thresholds from \SIrange{0}{100}{\percent}, into steps of \SI{1}{\percent}. A confidence threshold considers all the confidence rates bigger than or equal to the stated one. The confidence threshold that optimises the F1-score is selected for the model's normal operation.

We evaluated the four trained models to identify tomatoes using the test set. The whole inference process occurred on the Google Colab server, using a Tesla T4 GPU with a computation capability of \num{7.5}.

%%%%%%%%%%%%%%%%%%%%%%%%%%%%%%%%%%%%%%%%%%
\section{Results and Discussion}
\label{sec:results}

This section evaluates the four SSD models and the YOLO model to detect tomatoes in greenhouses. As mentioned in Section \ref{sec:training}, the trained models were evaluated using the metrics defined for the Pascal VOC challenge. Besides, some additional metrics were considered. In summary, we considered the following evaluation metrics:
\begin{itemize}
 \item Recall $\times$ precision curve;
 \item mAP (mean Average Precision);
 \item Total recall;
 \item Total precision;
 \item F1-score;
 \item Inference time.
\end{itemize}

As mentioned in Section \ref{sec:training}, before proceeding to evaluate the ANN's performance, the models required defining the best confidence threshold first. This value is the confidence threshold that maximises the F1-score \mbox{(Figure~{\ref{fig: f1 validation}})} because it find the best balance between the precision and recall, optimising the number of TPs while avoiding the FPs {and FNs \mbox{(Figure~{\ref{fig: validation TP-FP-FN}})}}. Figure~\ref{fig: f1 validation} reports the evolution of the F1-score with the variation of the confidence threshold for cross-validation. From this figure, we can quickly infer that some models have better behaviour than others. Models with flattened curves indicate higher confidence in their predictions and a low amount of FPs {and FNs}. Here, we can infer the maximum F1-score for each model and its confidence threshold (Table~\ref{tab: confthd vs f1}). These values are used to characterise the models for prediction purposes fully. 

Particular attention should be given to SSD MobileNet v2. This ANN model almost has no FPs {(Figure~{\ref{fig: validation TP-FP-FN}})}. This particularity is essential to avoid trying to harvest non-fruits and consequently damage the tree or the robot.

\begin{specialtable}[H]
\leavevmode
%\centering
\caption{Confidence threshold for each DL model that optimises the F1-score metric.}
\label{tab: confthd vs f1}
\begin{tabular}{l@{\hspace{2.4cm}}c@{\hspace{2.4cm}}c@{\hspace{2.4cm}}}
\toprule
\textbf{} & \textbf{Confidence $\geqslant$} & \textbf{F1-Score} \\ \midrule
YOLOv4 tiny & \SI{49}{\percent}     & \SI{85.92}{\percent} \\
SSD Inception v2 & \SI{21}{\percent}     & \SI{89.85}{\percent} \\
SSD MobileNet v2 & \SI{40}{\percent}     & \SI{82.22}{\percent} \\
SSD ResNet 50 & \SI{46}{\percent}     & \SI{90.46}{\percent} \\
SSD ResNet 101 & \SI{34}{\percent}     & \SI{81.75}{\percent} \\ \bottomrule
\end{tabular}
\end{specialtable}
\vspace{-20pt}

\begin{figure}[H]
 % \centering
 \includegraphics[width=\linewidth]{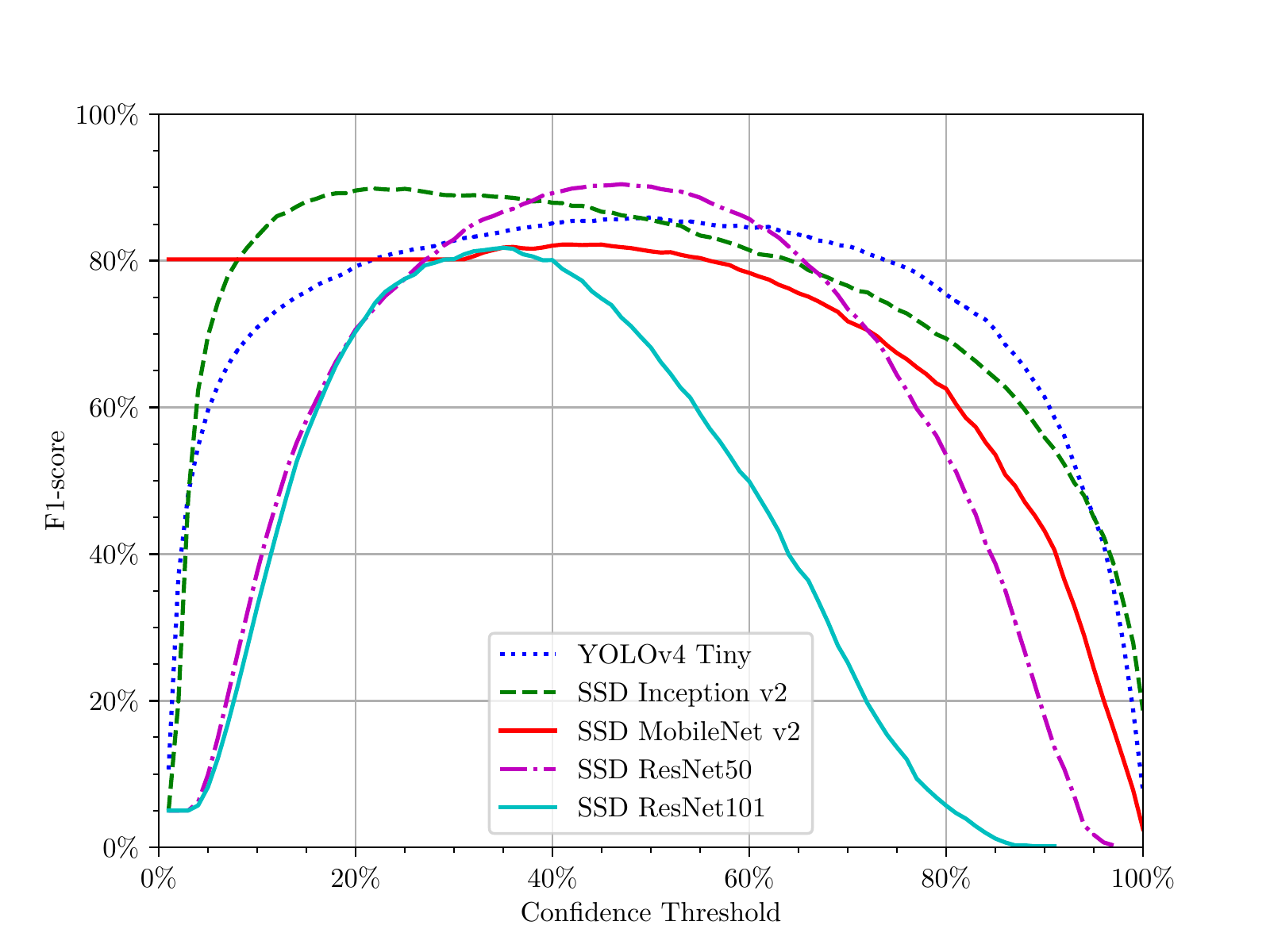}
 \caption{Evolution of the F1-score with the variation of the confidence threshold for all DL models in the validation set without augmentation.}
 \label{fig: f1 validation}
\end{figure}

\begin{figure}[H]
 % \centering
 \begin{subfigure}[b]{0.5\textwidth}
  \centering
  \includegraphics[width=\textwidth]{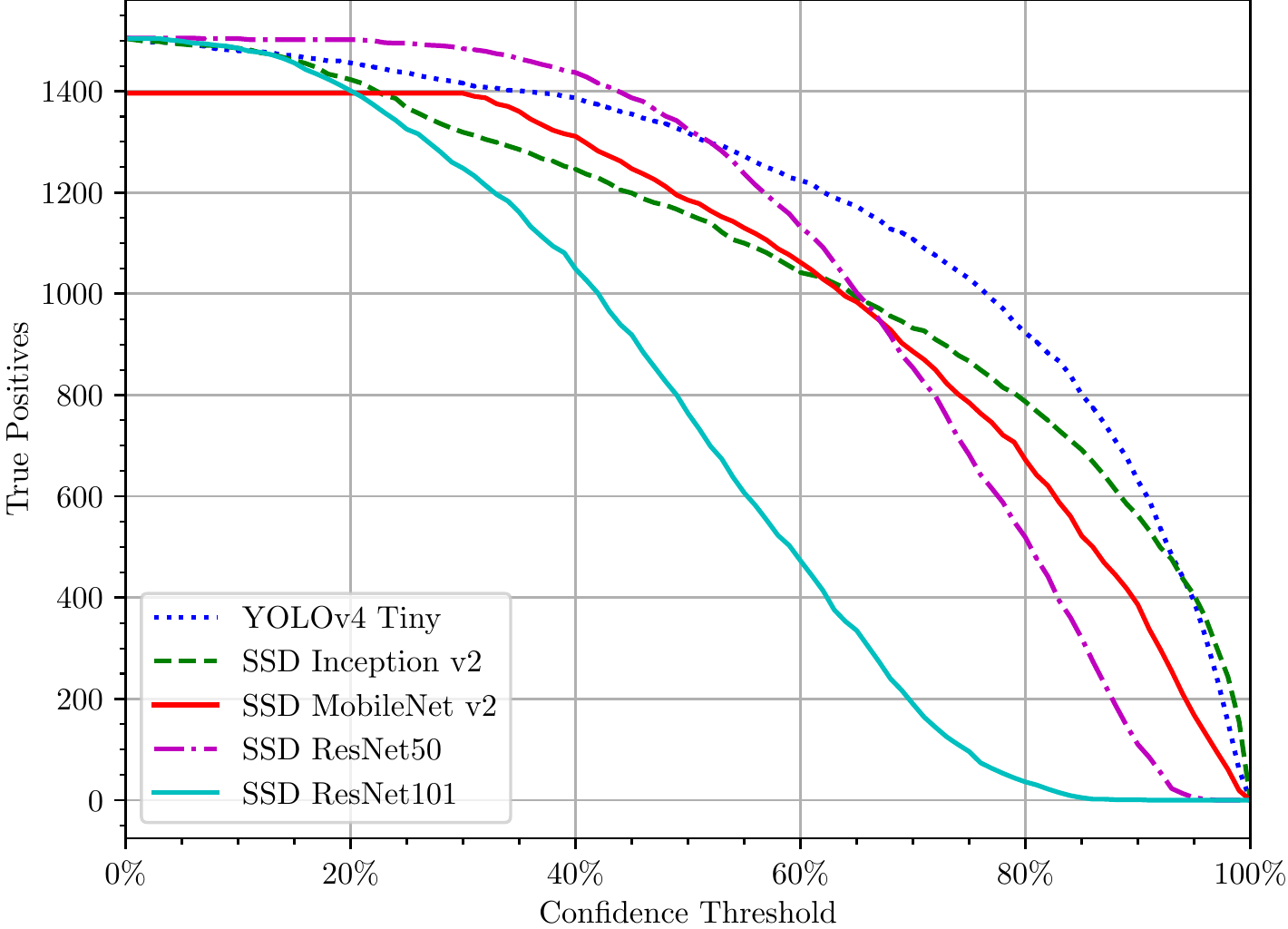}
  \caption{True Positives}
  \label{fig: val TP}
 \end{subfigure}%
 \hfill%
 \vspace{10pt}

 \begin{subfigure}[b]{0.5\textwidth}
  \centering
  \includegraphics[width=\textwidth]{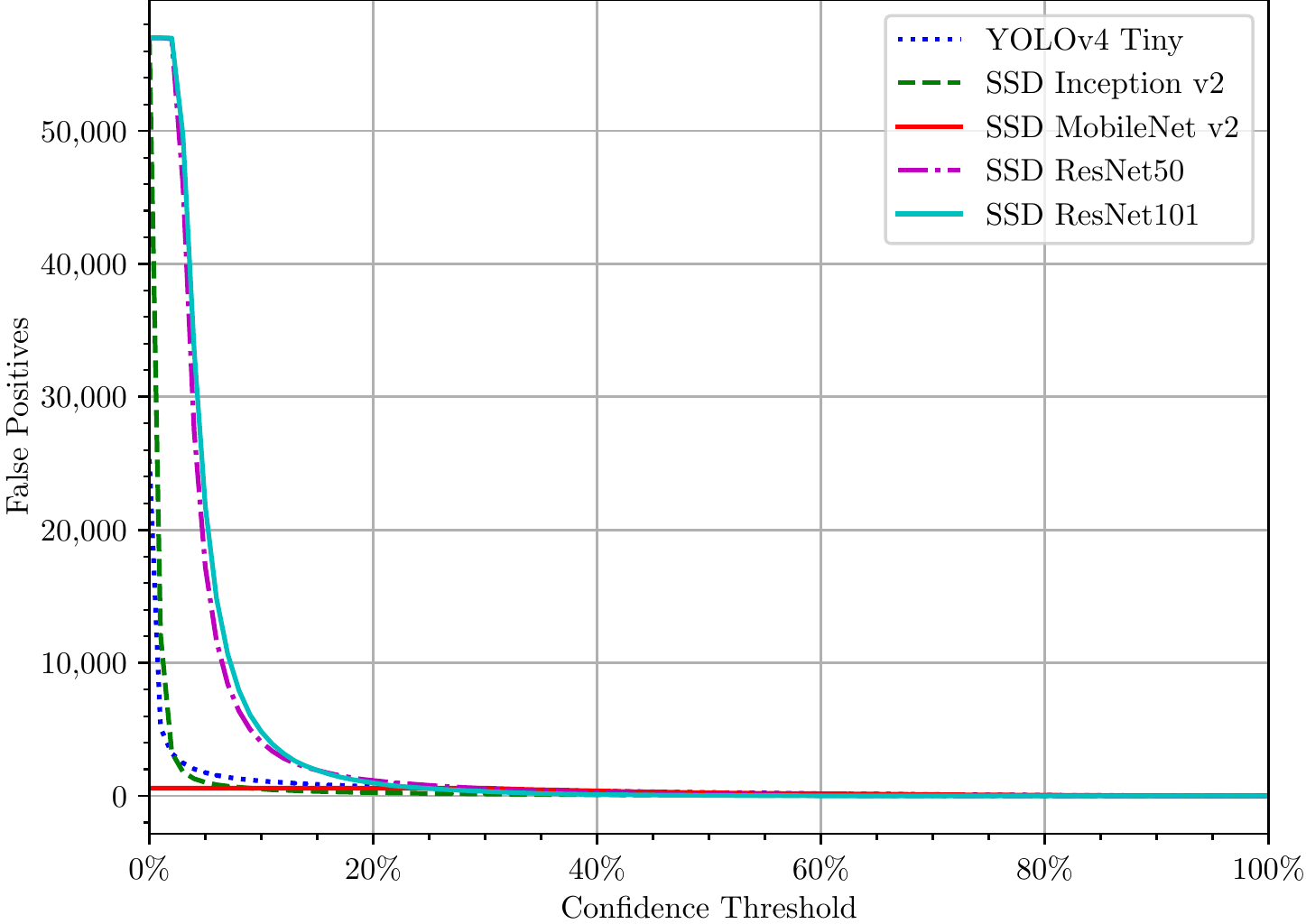}
  \caption{False Positives}
  \label{fig: val FP}
 \end{subfigure}
 
\vspace{10pt}%
 \begin{subfigure}[b]{0.5\textwidth}
  \centering
  \includegraphics[width=\textwidth]{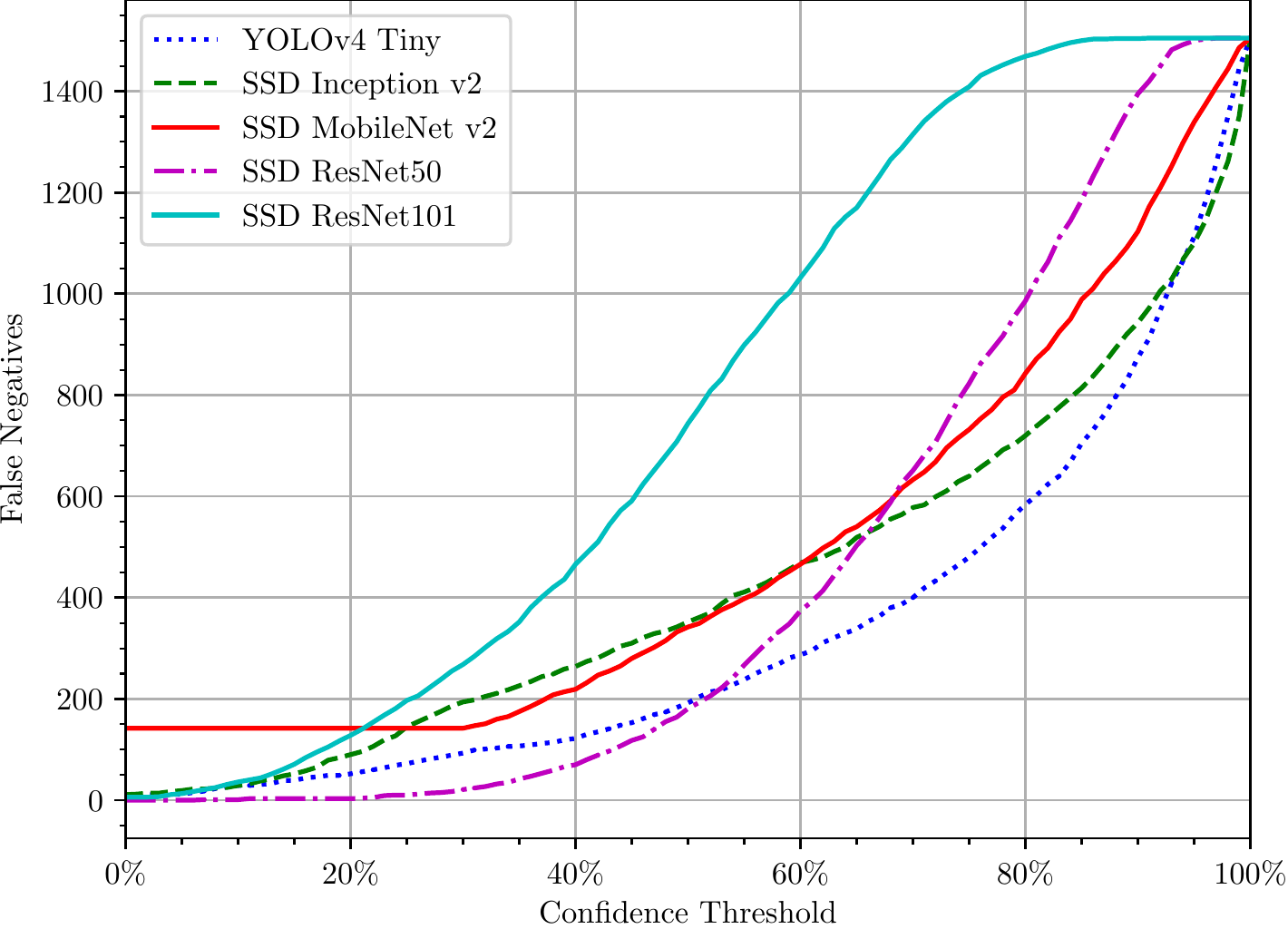}
  \caption{False Negatives}
  \label{fig: val FN}
 \end{subfigure}%
\vspace{10pt} 
 \caption{{Evolution of the number of TPs, FPs, and FNs with the increase of the confidence threshold.}}
 \label{fig: validation TP-FP-FN}%MDPI: Plot replaced accordingly
\end{figure}

While the previous analysis provided the benchmark in the validation set, the following performed the test set's performance analysis. This set was an independent collection of images and allowed understanding the generalisation capacity of the trained DL models. We started with a two-approach study and then converged to only the use of fully characterised models. First, we highlight the advantage of limiting the confidence rate. 

Using all the predictions for detecting tomatoes in images of a greenhouse's culture, a smooth precision $\times$ recall curve was built (Figure~\ref{fig:PRC_0}). This curve established the compromise between the recall rate and the precision rate, with the evolution of the prediction confidence score~\cite{padillaCITE2020}. Higher confidence rates tend to have higher precision in their predictions, but a lower recall rate. All the models, except SSD MobileNet v2, had near a \num{100}\si{\percent} recall rate, but in this stage, the precision was near \SI{0}{\percent}. The best performing model was the one with the highest Area Under the Curve (AUC)~\cite{padillaCITE2020}. Therefore, from \mbox{Figure~\ref{fig:PRC_0},} YOLOv4 Tiny and SSD ResNet 50 had similar results and were the best performing models. However, the low precision at higher recall rates and the lower total recall and F1-score (Table~\ref{tab:evaluation-results}) mean that the models have much prediction noise and many false positives (knowing that SSD ResNet50 has the worst results). Therefore, while considering all the model predictions, using the F1-score as a balanced metric between the recall and the precision, SSD MobileNet v2 was the best performing model. 

% start a new page without indent 4.6cm
%\clearpage
\end{paracol}
\nointerlineskip
\begin{specialtable}[H]
 \widetable
 \caption{{Results of the different SSD} and YOLO models over many metrics, considering all the predictions and the best computed confidence threshold.}
 % Is the bold number necessary in the table? If yes, please explain then in caption or mian text.
 %S.A.M: It's not need, but the idea is to highlight the best results.
 \label{tab:evaluation-results}
\begin{tabular}{l@{\hspace{2cm}}c@{\hspace{0.7cm}}c@{\hspace{0.7cm}}c@{\hspace{0.6cm}}c@{\hspace{0.6cm}}c@{\hspace{0.6cm}}c@{\hspace{0.6cm}}}
\toprule
\textbf{Model} & \textbf{Confidence $\geqslant$} & \textbf{Inference Time} & \textbf{mAP}   & \textbf{Precision}  & \textbf{Recall}   & \textbf{F1}    \\ \midrule
YOLOv4 Tiny & \SI{0}{\percent}   & {\SI{4.87}{\milli\second}} & {\SI{77.19}{\percent}} & \SI{6.38}{\percent}  & \SI{97.52}{\percent}  & \SI{11.98}{\percent}  \\
SSD Inception v2 & \SI{0}{\percent}   & \SI{24.75}{\milli\second} & \SI{70.39}{\percent}  & \SI{3.53}{\percent}  & \SI{95.82}{\percent}  & \SI{6.82}{\percent}  \\
SSD MobileNet v2 & \SI{0}{\percent}   & \SI{16.44}{\milli\second} & \SI{57.99}{\percent}  & {\SI{78.07}{\percent}} & \SI{62.44}{\percent}  & {\SI{69.39}{\percent}} \\
SSD ResNet50 & \SI{0}{\percent}   & \SI{47.78}{\milli\second} & \SI{75.74}{\percent}  & \SI{3.6}{\percent}  & {\SI{97.62}{\percent}} & \SI{6.94}{\percent}  \\
SSD ResNet101 & \SI{0}{\percent}   & \SI{59.78}{\milli\second} & \SI{66.88}{\percent}  & \SI{3.55}{\percent}  & \SI{96.32}{\percent}  & \SI{6.85}{\percent}  \\ \midrule
YOLOv4 Tiny & \SI{49}{\percent}   & {\SI{4.87}{\milli\second}} & \SI{47.48}{\percent}  & \SI{88.39}{\percent}  & \SI{49.33}{\percent}  & \SI{63.32}{\percent}  \\
SSD Inception v2 & \SI{21}{\percent}   & \SI{24.75}{\milli\second} & \SI{48.54}{\percent}  & \SI{85.31}{\percent}  & \SI{50.93}{\percent}  & \SI{63.78}{\percent}  \\
SSD MobileNet v2 & \SI{40}{\percent}   & \SI{16.44}{\milli\second} & {\SI{51.46}{\percent}} & \SI{84.37}{\percent}  & {\SI{54.40}{\percent}} & {\SI{66.15}{\percent}} \\
SSD ResNet50 & \SI{46}{\percent}   & \SI{47.78}{\milli\second} & \SI{42.62}{\percent}  & {\SI{92.51}{\percent}} & \SI{43.59}{\percent}  & \SI{59.26}{\percent}  \\
SSD ResNet101 & \SI{34}{\percent}   & \SI{59.78}{\milli\second} & \SI{36.32}{\percent}  & \SI{88.63}{\percent}  & \SI{38.13}{\percent}  & \SI{53.32}{\percent}  \\ \bottomrule
\end{tabular}
\end{specialtable}
\begin{paracol}{2}
%\linenumbers
\switchcolumn
\vspace{-20pt}

\begin{figure}[H]
 % \centering
 \includegraphics[width=0.7\linewidth]{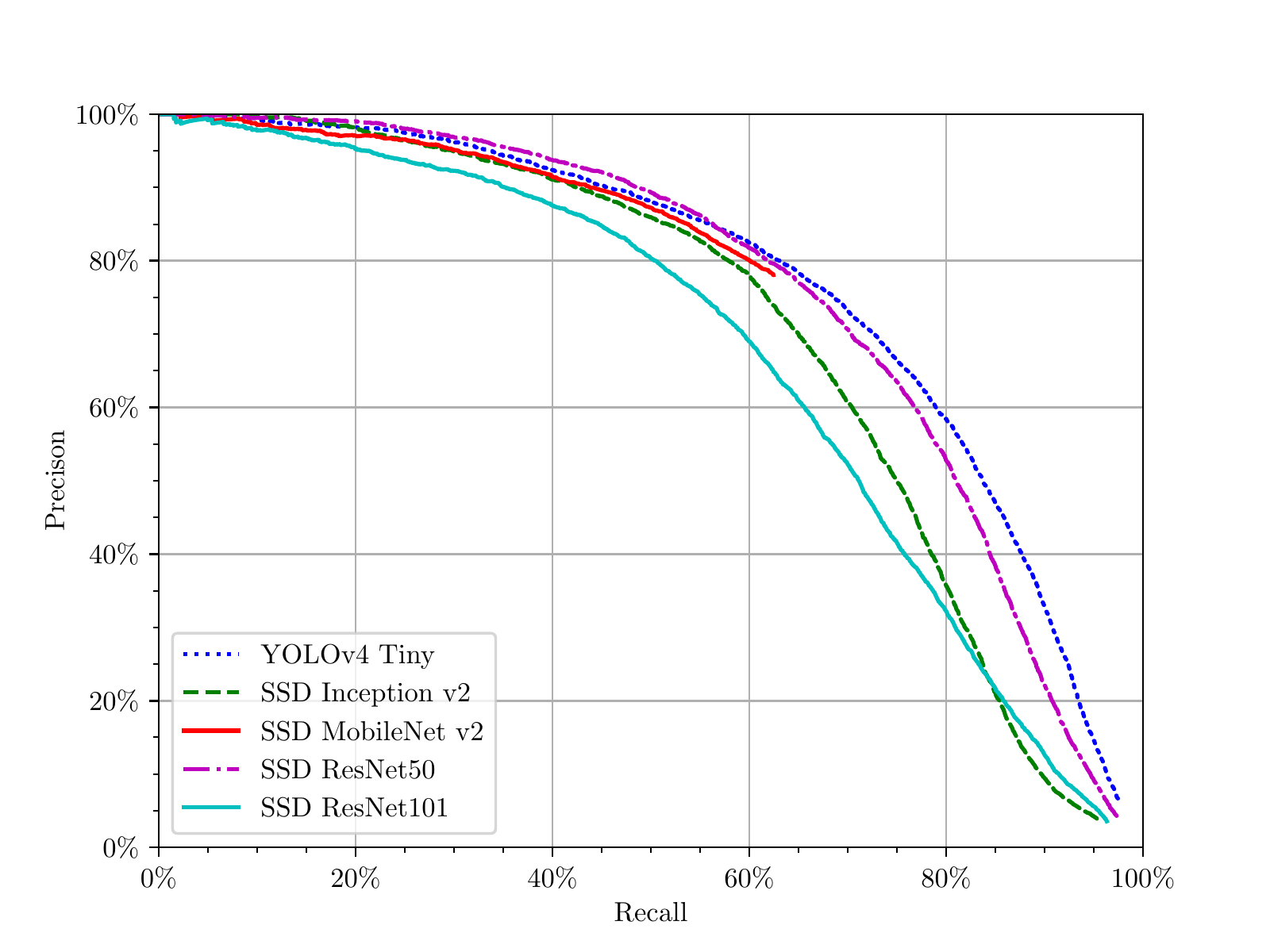}
 \caption{Precision $\times$ recall curve in the test set considering all the predictions.}
 \label{fig:PRC_0}
\end{figure}

Most of the time, the low precision rates were due to the low confidence of predictions, as proven in the confidence threshold tuning process. If we performed an additional filtering process on the predictions, considering the best computed confidence threshold to maximise the F1-score in the validation set (Table~\ref{tab: confthd vs f1} and Figure~\ref{fig: f1 validation}), the precision increased (\mbox{Table~\ref{tab:evaluation-results}}). Doing this, also the precision $\times$ recall curve (Figure~\ref{fig:PRC_calibrated}) was transformed through a truncation process. All the predictions had a precision rate higher than \SI{80}{\percent}, but a recall rate lower than \SI{60}{\percent}, as illustrated in Figure~\ref{fig:PRC_calibrated}. For the fully characterised models, SSD MobileNet v2 continued to be the best performing model. However, despite the slightly lower results for real-time purposes, YOLOv4 Tiny seemed to have an interesting inference time. A particular note should be realised for this model. It is a quantised model (int8), while the others are not (float32). Therefore, a better analysis should be done to compare the results of all the models quantised. Finally, it was easier to conclude that SSD ResNet101 was a complex model for this problem and overfit, performing worst. 
\vspace{-13pt}

\begin{figure}[H]
 %\centering
 \includegraphics[width=0.7\linewidth]{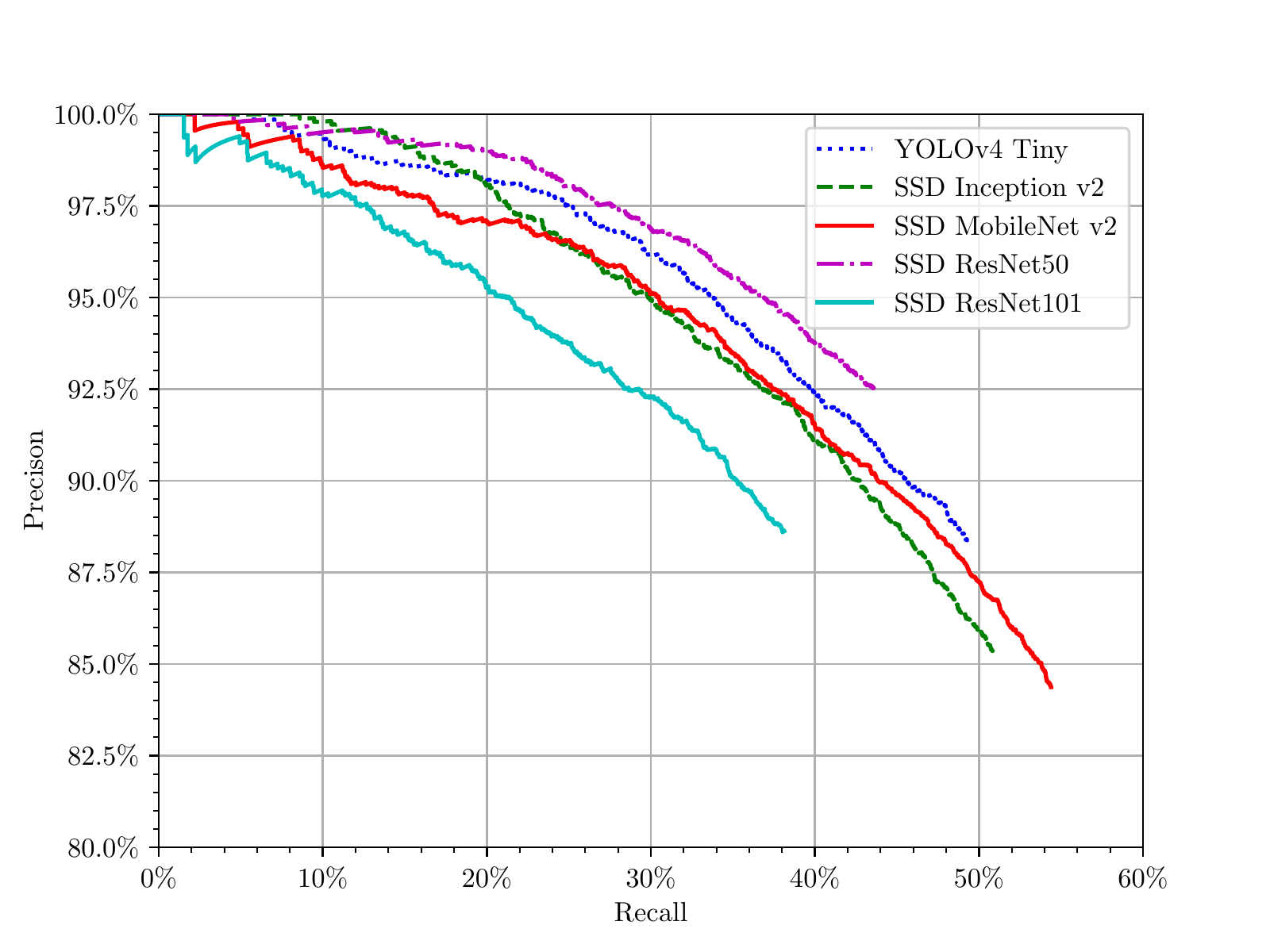}
 \caption{Precision $\times$ recall curve in the test set using the calibrated confidence threshold.}
 \label{fig:PRC_calibrated}
\end{figure}

{Quantisation is the process of mapping the continuous or large sets (here in the scope of 32 bit float) to a restricted set. In deep learning, it is common to map this number from the weights and bias of the neurons into 8 bit integers. The way edge devices compute makes this kind of ANN more suitable to run in real time. Typically, the performance of the ANN is irrelevantly penalised, but it becomes faster~{\cite{wu2020}}.}

In general, all the illustrated models were not generic enough to successfully characterise the class tomato to detect all the tomatoes. The results became much worse between the validation set and the test set. Therefore, the amount and the variability of data should be increased.

From Figure~\ref{fig:comp-results}, it is easy to verify that using all the predictions from the inference process resulted in many false positives. Using filtered results by a threshold or similar process was significant for all models, except SSD MobileNet v2. This model, as demonstrated in Figures~\ref{fig: f1 validation} and \ref{fig:PRC_0}, was well balanced between precision and recall and had a high confidence rate in its predictions, never reaching the situation of near \SI{0}{\percent} precision, i.e., any prediction was wrong. Additionally, this model also ensured a precision rate higher than \SI{80}{\percent}. Therefore, SSD MobileNet v2 can be used without any filtering process without compromising the results. 

For better understanding the capability of the different models, we performed additional analysis of the results, considering representative images from the dataset for specific situations:
\begin{enumerate}[label=(\roman*)]
 \item darkened tomatoes;
 \item occluded tomatoes;
 \item overlapped tomatoes.
\end{enumerate}

\begin{figure}[H]
 \centering
 \begin{subfigure}[b]{0.19\textwidth}
  \centering
  \includegraphics[width=\textwidth]{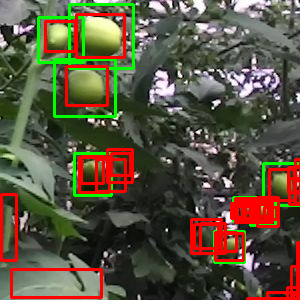}
  \caption{YOLOv4 Tiny}
  \label{fig:comp-yolov4-100}
 \end{subfigure}%
 \hfill%
 \begin{subfigure}[b]{0.19\textwidth}
  \centering
  \includegraphics[width=\textwidth]{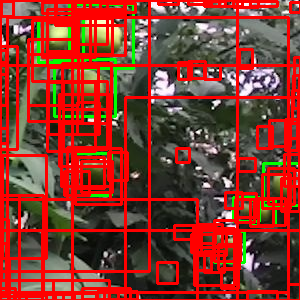}
  \caption{SSD Inception v2}
  \label{fig:comp-inception-100}
 \end{subfigure}%
 \hfill%
 \begin{subfigure}[b]{0.19\textwidth}
  \centering
  \includegraphics[width=\textwidth]{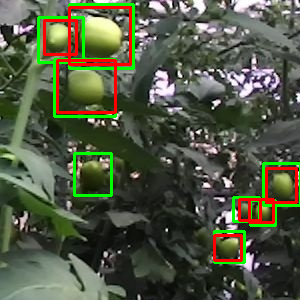}
  \caption{SSD MobileNet v2}
  \label{fig:comp-mobilenet-100}
 \end{subfigure}\\%
 \hfill%
 \vspace{10pt}

 \begin{subfigure}[b]{0.19\textwidth}
  \centering
  \includegraphics[width=\textwidth]{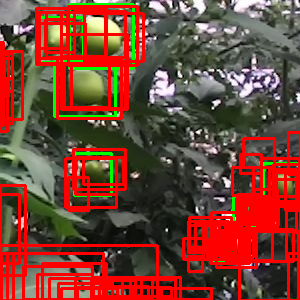}
  \caption{SSD ResNet 50}
  \label{fig:comp-resnet50-100}
 \end{subfigure}%
 \hfill%
 \begin{subfigure}[b]{0.19\textwidth}
  \centering
  \includegraphics[width=\textwidth]{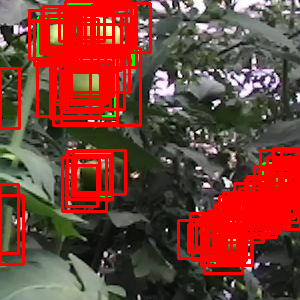}
  \caption{SSD ResNet 101}
  \label{fig:comp-resnet101-100}
 \end{subfigure}%
 \hfill%
\vspace{10pt} 
 \begin{subfigure}[b]{0.19\textwidth}
  \centering
  \includegraphics[width=\textwidth]{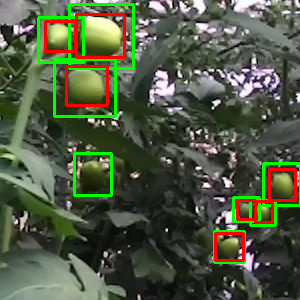}
  \caption{YOLOv4 Tiny}
  \label{fig:comp-yolov4}
 \end{subfigure}%
 \hfill%
 \begin{subfigure}[b]{0.19\textwidth}
  \centering
  \includegraphics[width=\textwidth]{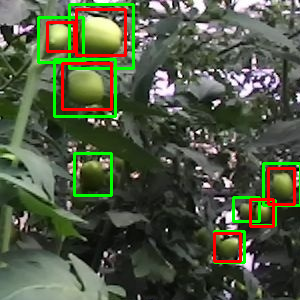}
  \caption{SSD Inception v2}
  \label{fig:comp-inception}
 \end{subfigure}%
 \hfill%
 \begin{subfigure}[b]{0.19\textwidth}
  \centering
  \includegraphics[width=\textwidth]{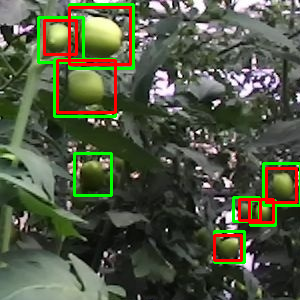}
  \caption{SSD MobileNet v2}
  \label{fig:comp-mobilenet}
 \end{subfigure}%
 \hfill%
 \vspace{10pt}
 \begin{subfigure}[b]{0.19\textwidth}
  \centering
  \includegraphics[width=\textwidth]{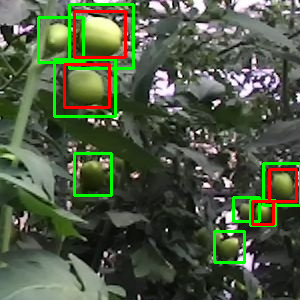}
  \caption{SSD ResNet 50}
  \label{fig:comp-resnet50}
 \end{subfigure}%
 \hfill%
 \begin{subfigure}[b]{0.19\textwidth}
  \centering
  \includegraphics[width=\textwidth]{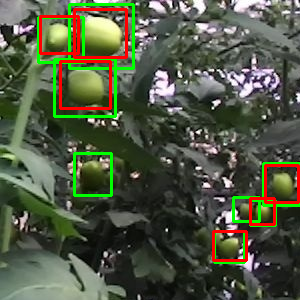}
  \caption{SSD ResNet 101}
  \label{fig:comp-resnet101}
 \end{subfigure}%
 \vspace{6pt}
 \caption{Comparison between using unfiltered images (\textbf{a}--\textbf{e}) and filtered images through the computed confidence threshold (\textbf{f}--\textbf{j}).}
 \label{fig:comp-results}
\end{figure}

Figure~\ref{fig:dark-results} is a representative result of darkening tomatoes, which happens while the robot enters the greenhouse or when sun-protected in the shadow of other plants or leaves. For the current situation, all models had similar results. However, SSD MobileNet v2 performed slightly better, detecting one more tomato.

\begin{figure}
 \centering
 \begin{subfigure}[b]{0.19\textwidth}
  \centering
  \includegraphics[width=\textwidth]{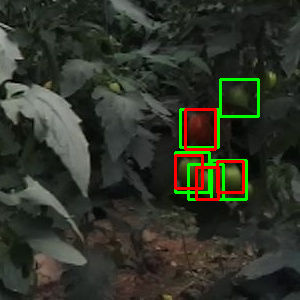}
  \caption{YOLOv4 Tiny}
  \label{fig:dark-yolov4}
 \end{subfigure}%
 \hfill%
 \begin{subfigure}[b]{0.19\textwidth}
  \centering
  \includegraphics[width=\textwidth]{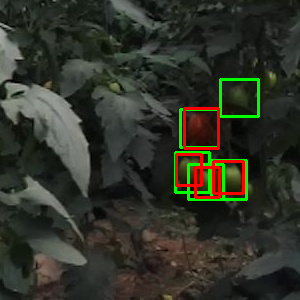}
  \caption{SSD Inception v2}
  \label{fig:dark-inception}
 \end{subfigure}%
 \hfill%
 \begin{subfigure}[b]{0.19\textwidth}
  \centering
  \includegraphics[width=\textwidth]{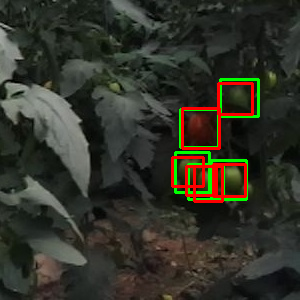}
  \caption{SSD MobileNet v2}
  \label{fig:dark-mobilenet}
 \end{subfigure}\\
 \hfill%
 \vspace{10pt}

 \begin{subfigure}[b]{0.19\textwidth}
  \centering
  \includegraphics[width=\textwidth]{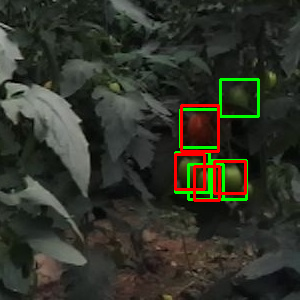}
  \caption{SSD ResNet 50}
  \label{fig:dark-resnet50}
 \end{subfigure}%
 \hfill%
 \begin{subfigure}[b]{0.19\textwidth}
  \centering
  \includegraphics[width=\textwidth]{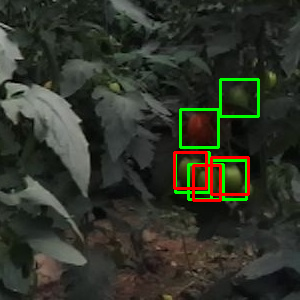}
  \caption{SSD ResNet 101}
  \label{fig:dark-resnet101}
 \end{subfigure}
  \vspace{6pt}
 \caption{Result comparison for darkened images.}
 \label{fig:dark-results}
\end{figure}

Occlusion refers to situations where a tomato is not fully visible. In these cases, tomatoes can be occluded by branches, stems, leaves or other tomatoes. Overlapping or clustering is a particular situation where other tomatoes occlude a tomato, and the detection system should detect both tomatoes. Figure~\ref{fig:occluded-results} demonstrates a typical situation of occlusion by leaves. For this situation, SSD MobileNet v2 had the best generalisation of the network, detecting tomatoes that have less than \SI{50}{\percent} of their area occluded. All other networks did not detect the occluded tomatoes.

\begin{figure}[H]
 \centering
 \begin{subfigure}[b]{0.19\textwidth}
  \centering
  \includegraphics[width=\textwidth]{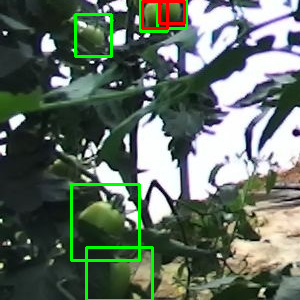}
  \caption{YOLOv4 Tiny}
  \label{fig:occluded-yolov4}
 \end{subfigure}
 \hfill
 \begin{subfigure}[b]{0.19\textwidth}
  \centering
  \includegraphics[width=\textwidth]{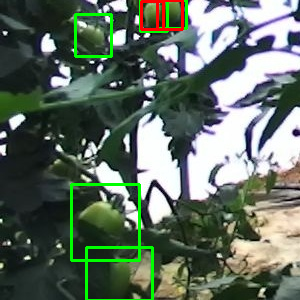}
  \caption{SSD Inception v2}
  \label{fig:occluded-inception}
 \end{subfigure}
 \hfill
 \begin{subfigure}[b]{0.19\textwidth}
  \centering
  \includegraphics[width=\textwidth]{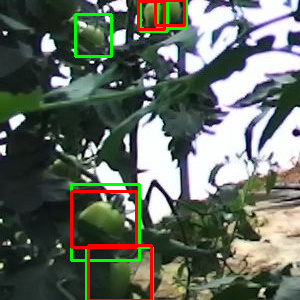}
  \caption{SSD MobileNet v2}
  \label{fig:occluded-mobilenet}
 \end{subfigure}\\%
 \hfill%
 \vspace{6pt}

 \begin{subfigure}[b]{0.19\textwidth}
  \centering
  \includegraphics[width=\textwidth]{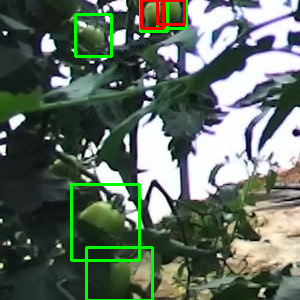}
  \caption{SSD ResNet 50}
  \label{fig:occluded-resnet50}
 \end{subfigure}
 \hfill
 \begin{subfigure}[b]{0.19\textwidth}
  \centering
  \includegraphics[width=\textwidth]{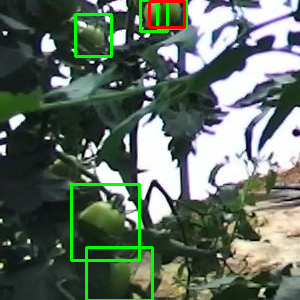}
  \caption{SSD ResNet 101}
  \label{fig:occluded-resnet101}
 \end{subfigure}
 \vspace{6pt}

 \caption{Result comparison for occluded tomatoes.}
 \label{fig:occluded-results}
\end{figure}

Considering the case of clusters of tomatoes or overlapped tomatoes {(Figure \ref{fig:overlap-results})}, all the DL models performed similarly. Therefore, all of them could be used equally for this situation. 

In summary, SSD MobileNet v2 was the best performing model. It could handle all the situations, avoiding false positives. Besides, SSD MobileNet v2 was also the fastest network among the SSD models, inferring in \SI{16}{\milli\second} with a high-performance GPU. However, we cannot ignore the performance of YOLOv4. YOLOv4 Tiny was faster than the others because it is an quantised model, which significantly reduced the processing time.

{In this work, we presented a real challenge, detecting tomato in the early ripening stage, where the colour feature was not so relevant, as stated in the literature review. We made this dataset public to support other works, and we analysed the most promising ANN models for edge computing. In these ANN models, most of the time, the confidence threshold was ignored or not tuned using clear criteria. We analysed this parameter, and we found there are ANN models where the performance can be significantly improved by tuning this parameter. With this work, we are able to move to the next stage, that is the deployment of these models into real robots and perception systems and benchmark against human labour---in terms of detection time, reliability and accuracy.}

\begin{figure}[H]
 \centering
 \begin{subfigure}[b]{0.19\textwidth}
  \centering
  \includegraphics[width=\textwidth]{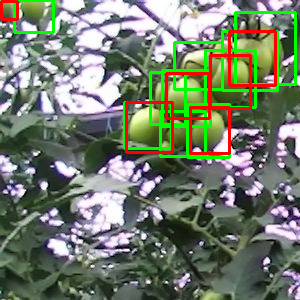}
  \caption{YOLOv4 Tiny}
  \label{fig:overlap-yolov4}
 \end{subfigure}
 \hfill
 \begin{subfigure}[b]{0.19\textwidth}
  \centering
  \includegraphics[width=\textwidth]{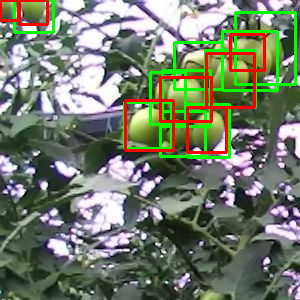}
  \caption{SSD Inception v2}
  \label{fig:overlap-inception}
 \end{subfigure}
 \hfill
 \begin{subfigure}[b]{0.19\textwidth}
  \centering
  \includegraphics[width=\textwidth]{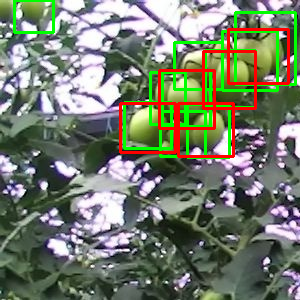}
  \caption{SSD MobileNet v2}
  \label{fig:overlap-mobilenet}
 \end{subfigure}\\
  \hfill
  \vspace{6pt}
  
 \begin{subfigure}[b]{0.19\textwidth}
  \centering
  \includegraphics[width=\textwidth]{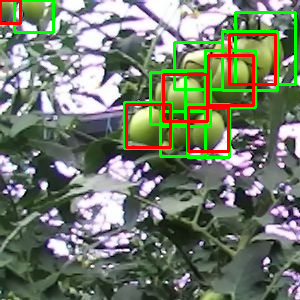}
  \caption{SSD ResNet 50}
  \label{fig:overlap-resnet50}
 \end{subfigure}
 \hfill
 \begin{subfigure}[b]{0.19\textwidth}
  \centering
  \includegraphics[width=\textwidth]{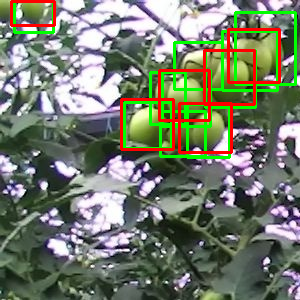}
  \caption{SSD ResNet 101}
  \label{fig:overlap-resnet101}
 \end{subfigure}
  \vspace{6pt}
 \caption{{Result comparison for overlapped tomatoes}.}%MDPI: it's not mentioned in the paper, please add %S.A.M: I am sorry, I forgot to mention
 \label{fig:overlap-results}
\end{figure}

%%%%%%%%%%%%%%%%%%%%%%%%%%%%%%%%%%%%%%%%%%

%%%%%%%%%%%%%%%%%%%%%%%%%%%%%%%%%%%%%%%%%%
\section{Conclusions}
\label{sec:conclusion}

The choice of SSD architectures under other DL models is based on whether they can infer images quickly in TPUs. YOLO models are also the SoA, and researchers have made their tiny models compatible with TPUs. We benchmarked four pre-trained SSD models from the TensorFlow database for detecting tomatoes and YOLOv4 Tiny from the Darknet database. The dataset was acquired inside the tomatoes' greenhouse using a stereo camera. The dataset was made publicly available at INESC~TEC {Research Data Repository} ({see} \url{https://rdm.inesctec.pt/} and \url{https://doi.org/10.25747/pc1e-nk92}{, updated on 14 May 2021})~\cite{dataset}.

The results proved that SSD MobileNet v2 was the best generalised and best performing model. Besides, it had the lowest ratio of FPs and performed the computation quickly. Although, if the inferring time was not fast enough, YOLOv4 Tiny also had impressive results and could process an image in about \SI{5}{\milli\second}.

The worst performing DL model was SSD ResNet 101, with an F1-score of \SI{38.13}{\percent}, and inferring images in 60\si{\milli\second}. SSD ResNet 101 is a complex model, so it needed more time to infer images. This complexity of the model favoured overfitting when the data were not enough or not representative of the class. Therefore, future work for this particular model is needed: 
\begin{enumerate}[label=(\roman*)]
 \item increasing the representativeness and the size of the dataset;
 \item adding regularisation to the model, penalising complex models.
\end{enumerate}

The additional future work will focus on:
\begin{enumerate}
 \item {creating new sub-classes to consider different tomatoes' contexts, as occluded tomatoes or darkened tomatoes};
 \item evaluating the performance of the detection system in on-time conditions inside the greenhouses;
 \item adding the capability to distinguish and evaluate the ripeness of tomatoes for harvesting procedures.
\end{enumerate}

%%%%%%%%%%%%%%%%%%%%%%%%%%%%%%%%%%%%%%%%%%
\vspace{6pt} 

%%%%%%%%%%%%%%%%%%%%%%%%%%%%%%%%%%%%%%%%%%
%% optional
%\supplementary{The following are available online at \linksupplementary{s1}, Figure S1: title, Table S1: title, Video S1: title.}


%%%%%%%%%%%%%%%%%%%%%%%%%%%%%%%%%%%%%%%%%%
\authorcontributions{Conceptualization, S.M. and F.N.d.S.; data curation, S.M., L.C. {and G.M.}; funding acquisition, F.N.d.S.; investigation, S.M. and L.C.; methodology, S.M. ; project administration, F.N.d.S.; software, S.M.; supervision, F.N.d.S., M.C., J.D. and A.P.M.; validation, F.N.d.S., J.D. and A.P.M.; visualization, S.M.; writing---original draft, S.M., L.C. and G.M.; writing---review and editing, F.N.d.S., M.C., J.D. and A.P.M. All authors read and agreed to the published version of the manuscript.}

%%%%%%%%%%%%%%%%%%%%%%%%%%%%%%%%%%%%%%%%%%
\funding{{The research leading to these results received funding from the European Union's Horizon 2020 {--} The EU Framework Programme for Research and Innovation 2014-2020, under Grant Agreement No. 857202 {--} DEMETER}. This work is also financed by National Funds through the Portuguese funding agency, FCT (Fundação para a Ciência e a Tecnologia) and co-financed by EFS (European Social Fund), within the scholarship SFRH/BD/147117/2019.} %Add this funding to the submission form

\institutionalreview{{Not applicable.}}
%mdpi: In this section, please add the Institutional Review Board Statement and approval number for studies involving humans or animals. Please note that the Editorial Office might ask you for further information. Please add ``The study was conducted according to the guidelines of the Declaration of Helsinki, and approved by the Institutional Review Board (or Ethics Committee) of NAME OF INSTITUTE (protocol code XXX and date of approval).'' OR ``Ethical review and approval were waived for this study, due to REASON (please provide a detailed justification).'' OR ``Not applicable'' for studies not involving humans or animals. You might also choose to exclude this statement if the study did not involve humans or animals.
%S.A.M.: This study does not involve animals or humans.

\informedconsent{{Not applicable.}}
%mdpi: Any research article describing a study involving humans should contain this statement. Please add ``Informed consent was obtained from all subjects involved in the study.'' OR ``Patient consent was waived due to REASON (please provide a detailed justification).'' OR ``Not applicable'' for studies not involving humans. You might also choose to exclude this statement if the study did not involve humans.Written informed consent for publication must be obtained from participating patients who can be identified (including by the patients themselves). Please state ``Written informed consent has been obtained from the patient(s) to publish this paper'' if applicable.
%S.A.M.: This study does not involve animals or humans.

\dataavailability{{The data presented in this study are openly available in INESC~TEC Research Data Repository at} {\href{https://doi.org/10.25747/PC1E-NK92}{DOI:10.25747/PC1E-NK92}}, reference number ii-2021-001.}
%mdpi: In this section, please provide details regarding where data supporting reported results can be found, including links to publicly archived datasets analyzed or generated during the study. Please refer to suggested Data Availability Statements in section ``MDPI Research Data Policies'' at https://www.mdpi.com/ethics. You might choose to exclude this statement if the study did not report any data.
%%%%%%%%%%%%%%%%%%%%%%%%%%%%%%%%%%%%%%%%%%
\acknowledgments{The authors would like to thank AMORINS and Mr. António through the COMPETE2020 program within the ROBOCARE project for making their tomato greenhouses available for data acquisition. The acquired data was used to generate the data set.}

%%%%%%%%%%%%%%%%%%%%%%%%%%%%%%%%%%%%%%%%%%
\conflictsofinterest{The authors declare no conflict of interest. The funders had no role in the design of the study; in the collection, analyses or interpretation of data; in the writing of the manuscript; nor in the decision to publish the results.} 

%%%%%%%%%%%%%%%%%%%%%%%%%%%%%%%%%%%%%%%%%%
%% Only for journal Encyclopedia
%\entrylink{The Link to this entry published on the encyclopedia platform.}

%%%%%%%%%%%%%%%%%%%%%%%%%%%%%%%%%%%%%%%%%%
%% Optional
\abbreviations{Abbreviations}{The following abbreviations are used in this manuscript: }

\noindent 
\begin{tabular}{@{}ll}
AI & Artificial Intelligence\\
ANN & Artificial Neural Network\\
COCO & Common Objects in Context\\
DP & Deep Learning\\
IMU & Inertial Moment Unit\\
GPU & Graphics Processing Unit\\
HSI & Hue, Saturation and Intensity\\
HSV & Hue, Saturation and Value\\
LiDAR & Light Detection and Ranging\\
mAP & mean Average Precision\\
ML & Machine Learning\\
RGB & Red, Green and Blue\\
RGB-D & Red, Green, Blue and Depth\\
ROS & Robotics Operating System\\
RVM & Relevance Vector Machine\\
SSD & Single-Shot MultiBox Detector\\
SVM & Support Vector Machine\\
TPU & TensorFlow Processing Unit\\
VRAM & Video Random-Access Memory\\
YOLO & You Only Look Once \\
\end{tabular}

%%%%%%%%%%%%%%%%%%%%%%%%%%%%%%%%%%%%%%%%%%
%% Optional
\appendixtitles{no} % Leave argument "no" if all appendix headings stay EMPTY (then no dot is printed after "Appendix A"). If the appendix sections contain a heading then change the argument to "yes".
\appendixstart
\appendix
\section{}
%\subsection{}
\label{ap:a}
%All appendix sections must be cited in the main text. In the appendixes, Figures, Tables, etc. should be labeled starting with `A', e.g., Figure A1, Figure A2, etc. 
\vspace{-6pt}

\begin{specialtable}[H]
 \caption{Model location in TensorFlow and Darknet databases. All SSD models are in the TensorFlow Models database at \url{http://download.tensorflow.org/models/object_detection/}<filename>. YOLOv4 Tiny is in the Darknet database at \url{https://github.com/AlexeyAB/darknet/releases/download/}<filename>.}
 \label{tab:ssd_models_location}
 \begin{tabular}{l@{\hspace{0.1cm}}c@{\hspace{0.15cm}}}
 \toprule
 \textbf{SSD Model} & \textbf{File Name} \\ \midrule
 SSD MobileNet v2 & ssd\_mobilenet\_v2\_coco\_2018\_03\_29.tar.gz \\
 SSD Inception v2 & ssd\_inception\_v2\_coco\_2018\_01\_28.tar.gz \\
 SSD ResNet 50 &{ssd\_resnet50\_v1\_fpn\_shared\_box\_predictor\_640x640\_coco14\_sync\_} \\
   &{2018\_07\_03.tar.gz} \\
 SSD ResNet 101 & {ssd\_resnet101\_v1\_fpn\_shared\_box\_predictor\_oid\_512x512\_sync\_} \\
  &2019\_01\_20.tar.gz \\
 YOLOv4 Tiny & darknet\_yolo\_v4\_pre/yolov4-tiny.conv.29 \\
 \bottomrule
 \end{tabular}

\end{specialtable}

\end{paracol}
%%%%%%%%%%%%%%%%%%%%%%%%%%%%%%%%%%%%%%%%%%
\reftitle{References}

\end{document}